\pdfoutput=1

\documentclass[11pt]{article}

\usepackage[]{EMNLP2023}

\usepackage{times}
\usepackage{latexsym}
\usepackage{microtype}
\usepackage{hyperref}
\usepackage{url}
\usepackage{booktabs}
\usepackage{amsmath}
\usepackage{graphicx}
\usepackage{colortbl}
\usepackage{xcolor}
\usepackage{tabularx}
\usepackage{tcolorbox}
\usepackage{amsfonts}
\usepackage{colortbl}
\usepackage{xcolor}
\usepackage{pifont}
\usepackage{multirow}
\usepackage[linesnumbered,ruled,vlined]{algorithm2e} 
\SetKwInput{KwInput}{Input}    
\SetKwInput{KwOutput}{Output}

\definecolor{lightgray2}{gray}{0.93}
\definecolor{headerblue}{RGB}{200, 220, 240}
\definecolor{debertashade}{gray}{0.95}

\definecolor{darkblue}{rgb}{0, 0, 0.5}
\hypersetup{colorlinks=true, citecolor=darkblue, linkcolor=darkblue, urlcolor=darkblue}

\usepackage[T1]{fontenc}

\usepackage[utf8]{inputenc}

\usepackage{microtype}

\usepackage{inconsolata}

\title{\texttt{ARC}: Argument Representation and Coverage Analysis for Zero-Shot Long Document Summarization with Instruction Following LLMs}

\author{Mohamed Elaraby, Diane Litman \\
        University of Pittsburgh \\ Pittsburgh, PA, USA \\ \texttt{\{mse30,dlitman\}@pitt.edu}}

\begin{document}
\maketitle
\begin{abstract}
We introduce \textbf{\underline{A}}rgument \textbf{\underline{R}}epresentation \textbf{\underline{C}}overage (\texttt{ARC}), a bottom-up evaluation framework that assesses how well summaries preserve  salient arguments, a  crucial issue in summarizing high-stakes domains such as law. \texttt{ARC} provides an interpretable lens by distinguishing between different information types to be covered and by separating omissions from factual errors.  
Using \texttt{ARC}, we evaluate summaries from eight open-weight LLMs in two domains where argument roles are central: \textit{long legal opinions} and \textit{scientific articles}. Our results show that while LLMs capture some salient roles, they frequently omit critical information, particularly when arguments are sparsely distributed across the input. Moreover, \texttt{ARC} uncovers systematic patterns—showing how context window positional bias and role-specific preferences shape argument coverage—providing actionable guidance for developing more complete and reliable summarization strategies.  
\end{abstract}

\section{Introduction}
\begin{figure}[ht]
\small
\includegraphics[scale=0.53]{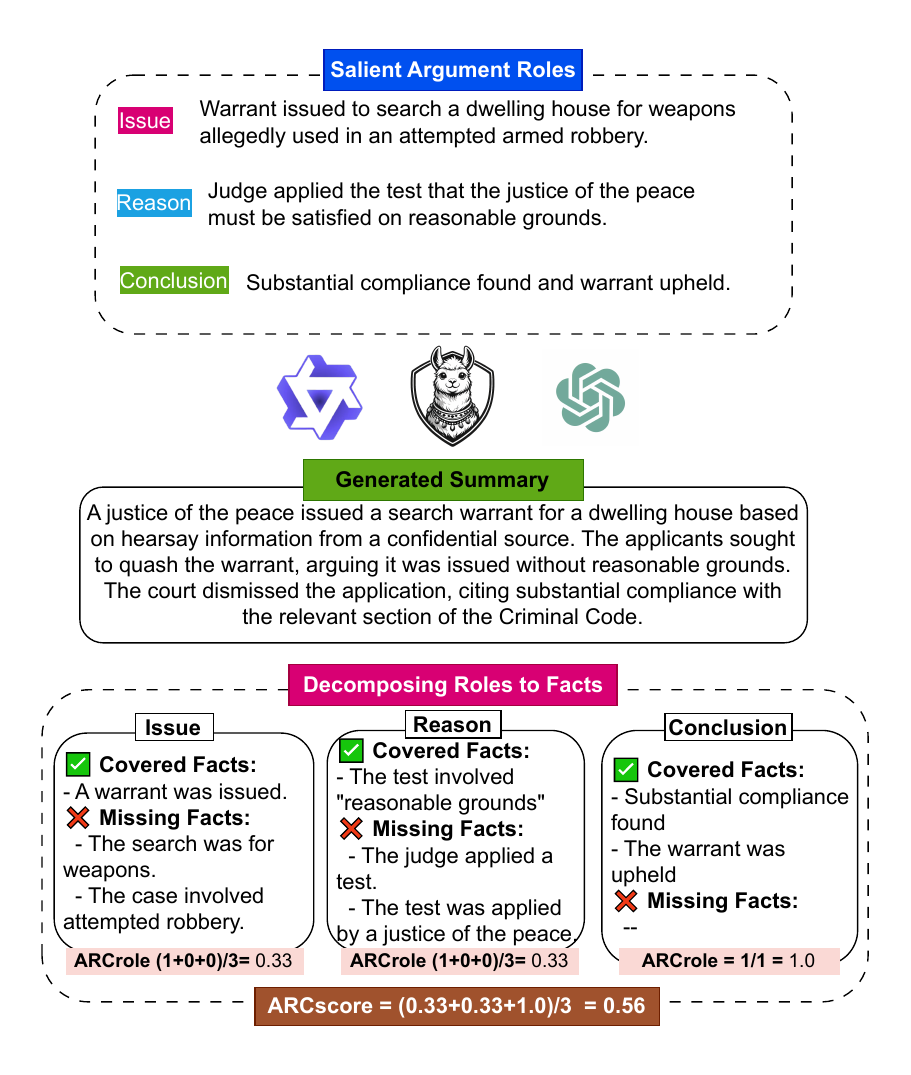} 
\caption{
Overview of the \texttt{ARC} framework, which computes coverage hierarchically from atomic facts to roles to an overall summary \texttt{ARC}\textsubscript{score}. Example shown from a \texttt{LlaMA3.18B} summary of a long legal opinion.}
\label{fig:arc_example}

\end{figure}

LLMs have achieved remarkable progress in abstractive summarization, producing summaries that are fluent, coherent, and often preferred by human evaluators in domains such as news~\cite{zhang-etal-2024-benchmarking,liu-etal-2024-learning}. This progress has shifted the focus of summarization research from \textit{how to generate fluent summaries} to \textit{how to properly evaluate them}. Evaluation is especially critical because, despite their fluency, LLMs frequently hallucinate or omit key content \cite{huang2025survey,kimfables}, undermining their reliability in high-stakes domains.  

In this work, we focus on the evaluation of long-context summaries generated by instruction-following LLMs, focusing on whether they preserve \textbf{salient argumentative content}. In crucial domains such as law, the most essential information is conveyed through \textit{argument roles} (the components of reasoning and main claims in the text). Preserving these roles is particularly challenging in long documents, where arguments are sparsely distributed across thousands of tokens~\cite{elaraby2022arglegalsumm}. The ability of LLMs to identify and retain argument roles is therefore a crucial test of their utility for reliable summarization in high-stakes settings.


\noindent\textbf{Why argument-based evaluation?}  
Prior work has shown that explicitly modeling argument roles can improve summarization quality~\cite{fabbri-etal-2021-convosumm,elaraby2022arglegalsumm,elaraby-etal-2023-towards}. Building on this, argument-based evaluation allows us to address a fundamental research question:  
\textbf{RQ1: Can instruction-following LLMs inherently identify and preserve argument roles without explicit supervision?}  
Additonally,  
we see two key benefits of argument-based evaluation where existing metrics fall short:  
$(1)$ \textit{interpreting which types of information are missing or misrepresented in generated summaries}, and  
$(2)$ \textit{guiding future alignment strategies for domain-specific summarization through a reproducible methodology}.  In this work, we focus on the first point by developing argument-based evaluation methods that make coverage interpretable.


 We introduce \textbf{\underline{A}}rgument \textbf{\underline{R}}epresentation \textbf{\underline{C}}overage (\texttt{ARC}), \footnote{The full codebase, along with instructions for obtaining the data, is publicly available at \url{https://github.com/EngSalem/ARCScore.git}.}
 a framework for evaluating how well LLM-generated summaries capture salient arguments.  
\texttt{ARC} leverages atomic fact decomposition of argument roles to measure coverage \textit{hierarchically}. As illustrated in Figure~\ref{fig:arc_example} (decomposing roles to facts), all salient arguments are first decomposed into atomic facts to assess \textbf{fact-level coverage}. Each atomic fact is then compared against the generated summary and labeled as either \textit{supported}, \textit{missing}, or \textit{contradicted}. These binary fact-level judgments are aggregated to compute \textbf{role-level coverage ($\text{\texttt{ARC}}_{\text{role}}$)}, providing fine-grained insight into how well each argument role is preserved. Finally, role-level scores are combined to produce \textbf{summary-level coverage ($\text{\texttt{ARC}}_{\text{score}}$)}, a holistic measure of how effectively a summary preserves salient argument roles.  
We apply \texttt{ARC} to study coverage in two domains where argument structure is essential for understanding the core of the document: \textit{long legal opinions} and \textit{scientific articles}.


Prior work shows that LLMs often exhibit positional biases—favoring content from the beginning or end of long documents~\cite{liu2024lost, ravaut-etal-2024-context, wan2024positional}. It is less clear, however, whether they also bias toward certain types of salient information that share the same structure (e.g., argument roles) during summarization. With \texttt{ARC}, we investigate additional two research questions:  \textbf{RQ2: How does the position of arguments in the source document affect their inclusion in summaries?} and \textbf{RQ3: Are certain argument roles disproportionately favored over others?} To answer the latter, we leverage \texttt{ARC}\textsubscript{role} to define a bias score that quantifies role-specific biases in saliency coverage.  

Our analysis with \texttt{ARC} across eight open-weight LLMs reveals:  
$(1)$ instruction-following models still struggle with saliency coverage, particularly when argument roles are sparsely distributed, underscoring the need for further alignment;  
$(2)$ positional bias in the context window negatively impacts coverage; and  
$(3)$ bias analysis shows that LLMs exhibit systematic preferences for certain argument roles, particularly in long legal opinions.


\begin{table*}[h]
\centering
\small
\resizebox{\textwidth}{!}{
\begin{tabular}{lccccc}
\hline
\textbf{Dataset} & \textbf{\# Docs} & \textbf{Input Length} & \textbf{Summary Length} & \textbf{\% Roles in Input} & \textbf{\% Roles in Summary} \\
\hline
\texttt{CANLII} & $1049$ & $122/4382/62786$ & $17/273/2072$ & $7.66\%$ & $66.51\%$ \\
\texttt{DRI} & $40$ & $3460/6505/11679$ & $67/221/298$ & $74.14\%$ & - \\
\hline
\end{tabular}
}
\caption{Statistics of the datasets, including the number of documents, input document length, reference summary length (min/mean/max words), and the percentage of argument roles in the input and summary. - indicates that value can't be directly computed from the corpus. \label{tab:dataset_stats}}

\end{table*}

\section{Related Work}

\noindent \textbf{Information Saliency in LLMs' Summaries.} 
Content selection remains a core challenge in summarization. \citet{trienes2025behavioral} found weak alignment between LLMs' saliency preferences and human judgments. While LLMs can produce summaries preferred over human references in domains like news \cite{zhang-etal-2024-benchmarking,liu-etal-2024-learning}, they still benefit from content planning. For example, \citet{adams-etal-2023-sparse} showed that planning entity mentions improves the information density in \texttt{GPT-4} generated  summaries at the same summary length when compared to summaries generated without entity planning. In similar veins, \citet{gantt-etal-2024-event,walden-etal-2025-cross} show that LLMs generate more focused and complete event-centered summaries when generation is explicitly conditioned on both the input document(s) and structured event representations.
 \textit{We extend this line of work by treating argument roles as a structured form of saliency and analyzing their preservation in LLM-generated summaries.} 

\noindent \textbf{LLMs in Long-Document Summarization.} 
LLMs face persistent issues when summarizing long texts, notably the \texttt{U}-shaped positional bias \cite{liu2024lost}—favoring content at the beginning and end while neglecting the middle \cite{ravaut-etal-2024-context}. This leads to degraded faithfulness in long-form outputs \cite{wan2024positional}. \textit{We expand this analysis by quantifying how positional bias affects the coverage of salient argumentative content.}

\noindent \textbf{Argument Mining in Abstractive Summarization.} 
Incorporating argument structures into summarization has shown promise across domains, including dialogues \cite{fabbri-etal-2021-convosumm}, legal texts \cite{xu2020using,xu2021toward,elaraby2022arglegalsumm, elaraby-etal-2023-towards} and scientific documents \cite{fisas2016multi}. \textit{We build on this by assessing whether instruction-following LLMs can cover salient arguments without the external argument role information}

\begin{table}[h]
\centering
\small
\renewcommand{\arraystretch}{1.2}
\setlength{\tabcolsep}{3pt}
\begin{tabular}{p{0.9\columnwidth}}
\toprule
\textbf{\texttt{CANLII} (Legal opinions)} \\
\midrule
\textcolor{blue}{\textbf{Issue:}} \textcolor{blue}{Damage to both vehicles exceeded the insurance deductibles and both parties claim damages against each other.} \\
\textcolor{red}{\textbf{Conclusion:}} \textcolor{red}{Fault for this accident was attributed $10\%$ to the defendant and $90\%$ to the plaintiff.} \\
\textcolor{teal}{\textbf{Reason:}} \textcolor{teal}{Jurisdictional error is not to be equated with error of law.} \\
\midrule
\textbf{\texttt{DRI} (Scientific documents)} \\
\midrule
\textcolor{purple}{\textbf{Own Claim:}} \textcolor{purple}{Semi-Lagrangian contouring offers an elegant and effective means for surface tracking with advantages over competing methods.} \\
\textcolor{orange}{\textbf{Background Claim:}} \textcolor{orange}{Accurate modeling of human motion remains a challenging task.} \\
\textcolor{gray}{\textbf{Data:}} \textcolor{gray}{Animation is constrained due to hardware constraints.} \\
\bottomrule
\end{tabular}
\caption{Examples of argument roles from \texttt{CANLII} and \texttt{DRI}. Colors distinguish different argument roles.}
\label{tab:argument_examples}
\end{table}
\noindent \textbf{Evaluating  Long-Form Summaries.} 
Standard metrics often fall short in reflecting human preferences, especially for long documents \cite{fabbri-etal-2021-summeval, krishna-etal-2023-longeval}. To improve reliability, recent work has introduced unit-based metrics—such as Atomic Content Units (ACUs) \cite{krishna-etal-2023-longeval} and structured factuality scores \cite{min-etal-2023-factscore, yang-etal-2024-fizz}—to reduce subjectivity in evaluation. \textit{Extending this idea, we propose \texttt{ARC}, which uses argument roles as evaluation units and introduces subatomic granularity to assess fine-grained argument coverage.}

\section{Datasets}
\label{sec:datasets}

We employ two datasets  that include both argument role annotations and reference summaries: \texttt{CANLII} \cite{xu2021toward}, representing the legal domain, and \texttt{DR. INVENTOR (DRI)} \cite{fisas2016multi}, representing the scientific domain. An overview of dataset statistics is presented in Table~\ref{tab:dataset_stats}. Both datasets consist of long-form documents paired with long-form reference summaries that average  $>150$ words \cite{krishna-etal-2023-longeval}.  More analysis and examples are given in Appendix~\ref{app:more_analysis}.

\subsection{Legal Opinions: \texttt{CANLII}}
The \texttt{CANLII} dataset consists of $1049$ legal cases annotated at the sentence level for argument roles.  Notably, only $7.66\%$ of the input text is labeled with argument roles, yet these argumentative sentences account for $66.51\%$ of the reference summaries (Table \ref{tab:dataset_stats}). This substantial mismatch highlights a haystack-like challenge: models must accurately identify and prioritize the sparse yet highly salient argumentative content when generating summaries.

Argument roles in \texttt{CANLII} are annotated using the \textbf{IRC} scheme \cite{xu2021toward}, which categorizes roles into three types:
 \textbf{Issues:} \textit{Legal questions raised in the case.}  \textbf{Reasons:} \textit{Justifications provided for judicial decisions. }\textbf{Conclusions:} \textit{Final rulings addressing the identified issues.} These annotations enable a fine-grained evaluation of argument coverage in generated summaries. Examples of IRCs are shown in Table \ref{tab:argument_examples}.

\subsection{Scientific Articles: \texttt{DRI}}
The \texttt{DRI} dataset consists of $40$ computer graphics articles, each annotated at the sentence level for $5$ rhetorical roles and paired with three human-written summaries. Notably, these rhetorical roles are not necessarily argumentative.
To address this limitation, the extended version of the dataset, \texttt{SCI-ARG} \cite{lauscher2018b}, enriches the \texttt{DRI} annotations by incorporating argument role annotations and their relations. These annotations  follow a modification of the $6$-argument roles described in \textit{Toulmin model} \cite{toulmin2003uses}, by reducing them into three types:  \textbf{Own Claim:} \textit{Sentences that directly support the author’s central argument.} \textbf{Background Claim:} \textit{Sentences that reference prior research or established domain knowledge.}  
\textbf{Data:} \textit{Empirical evidence that supports or refutes claims, such as experimental results or literature citations.} An example of each role is shown in Table \ref{tab:argument_examples}.  

Since the argument annotations are span-based, we map them back to complete sentences using lexical matching assigning the sentence with argument role spans if $>50\%$ of its words falls within the sentence boundaries.  A motivating feature behind selecting this corpus for our analysis is the sentence-level annotation for relevance scores on a Likert scale $(1-5)$, which indicates the degree of relevance to the summary. A relevance score of $4$ signifies that the sentence is "relevant to the summary," while a $5$ indicates it is "very relevant to the summary." In our evaluation, we focus on argument role coverage for sentences with argument roles and a Likert score of $5$ (indicating high relevant argument roles to the summary). Table \ref{tab:dataset_stats} shows that unlike legal opinions, where argument roles are sparsely distributed, scientific articles contain argumentative content throughout the document, posing a challenge of selectivity rather than retrieval. In \texttt{DRI}, sentences that contain at least 1 argument role account for $74.14\%$ of the input text (shown in Table \ref{tab:dataset_stats}). Although the dataset does not provide gold-standard summaries annotated for argument roles, we analyze the sentences with a Likert score of $5$ based on their argument role annotation. Among these sentences, $91.74\%$ contain at least one argument role, reinforcing the strong connection between argument roles and summarization relevance in this domain.


\section{The \texttt{ARC} Framework}
\label{sec:arc}

\subsection{Overview and Notations}

\noindent \textbf{Decomposing Roles to Facts.}  
For each argument role $r_i \in R$, where $R$ is the set of salient argument roles (roles that are essential to be included in the summary), we follow the decomposition algorithm of \citet{yang-etal-2024-fizz}. Specifically, we employ a strong LLM (\texttt{GPT4-o}) to decompose $r_i$ into a set of atomic facts $\mathbb{F}_{r} = \{f_1, f_2, \dots, f_{|\mathbb{F}_{r}|}\}$. 
These  facts are further filtered using an entailment check against $r_i$ to eliminate over-generated facts. \footnote{Decomposition algorithm in Appendix~\ref{app:decompose}.}  

\noindent \textbf{Role-Level Coverage.}  
Given a generated summary $S$, we define the indicator function:
\[
\resizebox{\columnwidth}{!}{$
\delta(f_i, S) = 
\begin{cases} 
1 & \text{if atomic fact } f_i \text{ is correctly covered in } S, \\[6pt]
0 & \text{otherwise}.
\end{cases}
$}
\]

The role-level coverage for $r$ is then:
\[
\texttt{ARC}_{\text{role}}(r, S) = \frac{1}{|\mathbb{F}_r|} \sum_{i=1}^{|\mathbb{F}_r|} \delta(f_i, S).
\]

\noindent \textbf{Summary-Level Coverage.}  
The overall coverage score, \texttt{ARC}\textsubscript{score}, across all roles is computed as aggregation of $\texttt{ARC}_{\text{role}}$ across $R$:
\[
\texttt{ARC}_{\text{score}}(S) = \frac{1}{|R|} \sum_{r \in R} \texttt{ARC}_{\text{role}}(r, S).
\]

\subsection{Evaluating Factual Units ($\delta$)}  
The main goal of the \texttt{ARC} framework is to provide an interpretable evaluation of coverage in generated summaries.  
To do so, the $\delta$ function distinguishes between two error types: \textit{missing facts} (information omitted) and \textit{incorrectly covered facts} (information misrepresented). \footnote{Examples of atomic facts are included in Appendix \ref{app:fact_errors}.}  
Although both reduce completeness, separating them reveals whether errors stem from failing to prioritize salient content or from factual mistakes during generation.  
We implement $\delta$ using an LLM judge. \footnote{See Appendix \ref{app:eval_prompts} for the evaluation prompt.}
We employ zero-shot LLMs, which have shown strong correlation with human judgments in summary evaluation \cite{liu2023g}.  
Given an atomic fact and a summary, we prompt the LLM to return a pair $(d, e)$, where $d \in \{0,1\}$ indicates whether the fact is supported (\textit{covered}) or not supported (\textit{missing or incorrect}), and $e$ provides a categorical interpretation of the error type \textit{(no error, missing, non-factual)}.  

\newcommand{\cmark}{\ding{51}}
\newcommand{\xmark}{\ding{55}}





\subsection{Benchmarking \texttt{ARC}\textsubscript{score}}  
To evaluate the effectiveness of \texttt{ARC}\textsubscript{score} in capturing holistic summary coverage, we benchmark it against existing automatic metrics using expert-annotated data.  

\noindent\textbf{Expert-Annotated Data.}  
Human-annotated datasets for fine-grained coverage evaluation are extremely scarce. To our knowledge, the only available resource is the dataset of \citet{elaraby-etal-2024-adding}, which includes $90$ legal opinion summaries annotated for salient argument coverage on a $4$-point Likert scale. \footnote{See Appendix~\ref{app:human_likert} for scale definitions.} Two legal experts independently annotated the data, achieving a moderate agreement of $\kappa=0.46$ (quadratic weighted).  
To improve the robustness of the annotations, we filtered out pairs with $> 1$ point of disagreement, resulting in $87$ article–summary pairs with improved agreement ($\kappa=0.605$). We use the expert average as the gold standard to compare against automatic metrics.


\noindent \textbf{Automatic  Metrics for Holistic Coverage.}  
We benchmark \texttt{ARC}$_\text{score}$ against a suite of strong automatic metrics.  
For all baselines, we treat the set of salient arguments as the \textit{hypothesis} and the generated summary as the \textit{premise/reference}.  
We first include the standard metrics reported in \citet{elaraby-etal-2024-adding}: \texttt{ROUGE-1, ROUGE-2, ROUGE-L}, and \texttt{BERTScore}.  
Next, we add entailment-based baselines from \citet{laban2022summac}, where \texttt{SummaC} computes entailment between hypothesis and reference at both sentence-level and document-level granularity.  We include both (zeroshot \texttt{zs} and convolution \texttt{conv} versions of \texttt{SummaC}).
Finally, we incorporate an LLM-judge baseline:  
\texttt{FactScore} \cite{min-etal-2023-factscore}, which—similar to \texttt{ARC}\textsubscript{score}—relies on decomposition for fact-level evaluation.  
We adapt \texttt{FactScore} for coverage by treating the generated summary as the knowledge base and the salient arguments as hypotheses to be tested.  
For fairness, both \texttt{ARC}\textsubscript{score} and \texttt{FactScore} use \texttt{GPT4-o} as the evaluator.  
We report correlation with expert averages using Kendall’s $\tau$ and Pearson’s $\rho$.  

\begin{table}[!t]
\small
\centering
\renewcommand{\arraystretch}{1.15}
\setlength{\tabcolsep}{4pt}
\resizebox{\columnwidth}{!}{%
\begin{tabular}{>{\bfseries}l|c c|c c}
\toprule
\multirow{2}{*}{\textbf{Metric}} & \multicolumn{2}{c|}{\textbf{Expert Avg.}} & \multicolumn{2}{c}{\textbf{Interpretability}} \\
\cmidrule(lr){2-5}
& $\tau$ & $\rho$ & Info Types & Errors \\
\midrule
\texttt{ROUGE-1} & $0.391$ & $0.539$ & \xmark & \xmark \\
\texttt{ROUGE-2} & $0.336$ & $0.475$ & \xmark & \xmark \\
\texttt{ROUGE-L} & $0.345$ & $0.479$ & \xmark & \xmark \\
\midrule
\texttt{BERTScore} & $0.354$ & $0.517$ & \xmark & \xmark \\
\midrule
\texttt{SummaC\textsubscript{ZS} (sent)} & $0.387$ & $0.537$ & \xmark & \xmark \\
\texttt{SummaC\textsubscript{ZS} (doc)} &  $0.375$ & $0.476$ & \xmark & \xmark \\
\texttt{SummaC\textsubscript{conv} (sent)} & $0.345$ & $0.459$ & \xmark & \xmark \\
\texttt{SummaC\textsubscript{conv} (doc)} & $0.352$ & $0.311$ & \xmark & \xmark \\
\midrule
\texttt{FactScore} (GPT4-o) & $0.405$ & $0.549$ & \xmark & \xmark \\
\midrule
ARC\textsubscript{score} (GPT4-o)   & $\mathbf{0.465}$ & $\mathbf{0.593}$ & \cmark & \cmark \\
\bottomrule
\end{tabular}%
}
\caption{
Metric correlations (Kendall’s $\tau$, Pearson’s $\rho$) with expert judgments using $87$ articles ($\kappa=0.605$). 
All rows: $p<0.05$.}
\label{tab:arc_benchmark_1col}
\end{table}

\begin{table}[!t]
\small
\centering
\renewcommand{\arraystretch}{1.1}
\setlength{\tabcolsep}{4pt}
\begin{tabular}{l c c}
\toprule
\textbf{Judge Model} & $\tau$ & $\rho$ \\
\midrule
\multicolumn{3}{l}{\textit{\bf{Proprietary models}}} \\
\underline{\texttt{GPT-4o}} & \underline{0.465} & \underline{0.593} \\
\texttt{GPT-4o-mini} & 0.446 & 0.582 \\
\midrule
\multicolumn{3}{l}{\textit{\bf{Open-weight  Instruction-following models}}} \\
\texttt{Qwen-2.5-7B-Instruct} & 0.329 & 0.480 \\
\texttt{Qwen-2.5-14B-Instruct} & 0.454 & 0.601 \\
\texttt{Llama-3.1-8B-Instruct} & \it{0.463} & \it{0.610} \\
\texttt{Mistral-8B-Instruct} & 0.443 & 0.599 \\
\midrule
\multicolumn{3}{l}{\textit{\bf{ Open-weight Reasoning models}}} \\
\texttt{QwQ-32B-AwQ} & 0.437 & 0.596 \\
\bf{\texttt{DeepSeek-R1-Distill-Qwen-14B}} & \bf{0.509} & \bf{0.638} \\
\bottomrule
\end{tabular}
\caption{ 
Expert-average correlations (Kendall’s $\tau$, Pearson’s $\rho$) for judges in ARC\textsubscript{score}. 
\textit{Italicized} = best instruction-following, \underline{underlined} = best proprietary, \textbf{bold} = overall best. $p<0.05$ for all rows.  
}
\label{tab:llm_judges}
\end{table}

Table~\ref{tab:arc_benchmark_1col} shows that among automatic metics, \texttt{ARC}\textsubscript{score} achieves the highest correlation with expert judgments,\footnote{See Appendix ~\ref{app:exp_corr} for correlations with each
expert.} outperforming standard evaluation metrics and LLM-based decomposition-based metrics such as \texttt{FactScore}.  
Crucially, \texttt{ARC}\textsubscript{score} improves the holistic view of coverage over baselines, while also providing interpretability that existing metrics lack.  
Unlike \texttt{FactScore}, which provides only fact-level coverage, \texttt{ARC} enables multi-level analysis:  
$(i)$ role-level coverage via \texttt{ARC}\textsubscript{role}, and  
$(ii)$ error-type analysis distinguishing between \textit{missing} and \textit{incorrect} facts.  
This decomposition allows us to diagnose not only how much coverage is achieved, but also which roles are underrepresented and why errors occur, making \texttt{ARC} a more effective diagnostic tool for granular coverage evaluation.  

\noindent \textbf{Alternative LLM-judges.}  
To make \texttt{ARC} more scalable and accessible, we explore alternatives to \texttt{GPT-4o} for fact-level evaluation that maintain high correlation with expert ratings while reducing inference cost and avoiding reliance on proprietary APIs that may be deprecated. We focus on two categories of open-weight models:  
$(i)$ \textit{instruction-following models.} We evaluate smaller open-weight models that fit within our compute and memory budgets, including \texttt{Qwen-2.5-Instruct} (7B and 14B), \texttt{Llama-3.1-8B-Instruct}, and \texttt{Mistral-8B-Instruct}.  
$(ii)$ \textit{Reasoning models.} We further examine open-weight reasoning models, specifically \texttt{QwQ-32B-AwQ} and \texttt{DeepSeek-R1-Distill-Qwen-14B}, which have demonstrated competitive performance with proprietary instruction-following and reasoning models on complex tasks, including verification.  
Table~\ref{tab:llm_judges} shows that \texttt{DeepSeek-R1-Distill-Qwen-14B} achieves the strongest overall correlation, surpassing even proprietary models such as \texttt{GPT-4o}.  
Among open-weight instruction-following models, \texttt{Llama-3.1-8B-Instruct} attains the highest correlation, outperforming \texttt{GPT-4o-mini}.  
We employ both proprietary (\texttt{GPT-4o}) and open-weight judge (\texttt{DeepSeek-R1-Distill-Qwen-14B}) to compute \texttt{ARC}\textsubscript{score}.  
To balance cost and scalability, \texttt{GPT-4o} is applied to a representative subset of $100$ \texttt{CANLII} (\texttt{CANLII}\textsubscript{100}) cases and the full \texttt{DRI} corpus, while \texttt{DeepSeek-R1-Distill-Qwen-14B} is used for both \texttt{CANLII}\textsubscript{100} and the full \texttt{CANLII} ($1049$ articles). Our goal is to  
 ensure that the sensitivity to the underlying judge (i.e., disagreements at the fact level) does not alter the overall coverage trends observed across models.

\begin{table}[t]
\centering
\resizebox{\columnwidth}{!}{
\begin{tabular}{l|cc|c|cc}
\toprule
\textbf{Model} & \multicolumn{2}{c|}{\texttt{CANLII\textsubscript{100}}} & \texttt{CANLII} & \multicolumn{2}{c}{\texttt{DRI}} \\
\cmidrule(lr){2-3}\cmidrule(lr){4-4}\cmidrule(lr){5-6}
 & GPT4-o & \cellcolor{gray!20}DeepSeek & \cellcolor{gray!20}DeepSeek & GPT4-o & \cellcolor{gray!20}DeepSeek \\
\midrule
\textbf{Reference}      & \bf{0.986} & \cellcolor{gray!10}\bf{0.977} & \cellcolor{gray!10}\bf{0.982} & \bf{0.838} & \cellcolor{gray!10}\bf{0.813} \\
Qwen-2.5-14B   & \it{0.677} & \cellcolor{gray!10}\it{0.707} & \cellcolor{gray!10}0.722 & 0.772 & \cellcolor{gray!10}0.797 \\
Qwen-2.5-7B    & 0.676 & \cellcolor{gray!10}0.691 & \cellcolor{gray!10}\it{0.725} & \it{0.799} & \cellcolor{gray!10} \it{0.806} \\
Qwen-2.5-3B    & 0.648 & \cellcolor{gray!10}0.702 & \cellcolor{gray!10}0.710 & 0.730 & \cellcolor{gray!10}0.759 \\
Qwen-2.5-1.5B  & 0.472 & \cellcolor{gray!10}0.512 & \cellcolor{gray!10}0.539 & 0.668 & \cellcolor{gray!10}0.628 \\
Mistral-8B     & 0.554 & \cellcolor{gray!10}0.609 & \cellcolor{gray!10}0.608 & 0.669 & \cellcolor{gray!10}0.697 \\
LLaMA-3.1-8B   & 0.641 & \cellcolor{gray!10}0.642 & \cellcolor{gray!10}0.678 & 0.749 & \cellcolor{gray!10}0.793 \\
LLaMA-3.2-3B   & 0.573 & \cellcolor{gray!10}0.603 & \cellcolor{gray!10}0.639 & 0.705 & \cellcolor{gray!10}0.740 \\
LLaMA-3.2-1B   & 0.451 & \cellcolor{gray!10}0.530 & \cellcolor{gray!10}0.545 & 0.674 & \cellcolor{gray!10}0.631 \\

\bottomrule
\end{tabular}}
\caption{Average \texttt{ARC}\textsubscript{score} across \texttt{CANLII\textsubscript{100}}, full \texttt{CANLII}, and \texttt{DRI}. 
Models are ordered by size. \textbf{Bold} indicates the human reference (upper bound), and \textit{italic} marks the best-performing model in each dataset.}
\label{tab:arc_scores}
\end{table}

\section{Analyzing Summary Coverage with \texttt{ARC}}
\label{sec:coverage_results}

In this section, we use \texttt{ARC} to evaluate summaries produced by long-context LLMs on the \texttt{CANLII} and \texttt{DRI} datasets.



\subsection{Obtaining Generated  Summaries}
\label{subsec:obtain_summs}



We evaluate eight long-context open-weight LLMs from the \texttt{LLaMA} \cite{grattafiori2024LlaMA}, \texttt{Mistral} \cite{jiang2024mixtral} and \texttt{Qwen} \cite{qwen2,qwen2.5}
families that meet our computational constraints. Specifically, we include \texttt{LLaMA-3.1-8B-Instruct}, \texttt{LLaMA-3.2-1B} and \texttt{3B}, \texttt{Qwen-2.5-Instruct-1.5B}, \texttt{3B}, \texttt{7B}, \texttt{14B}, and \texttt{Mistral-8B}. We vary model sizes to examine whether scaling improves coverage performance. Inference is conducted using \texttt{VLLM} \cite{kwon2023efficient} for efficiency and scalability, with \texttt{Qwen} models further extended to $130k$ tokens via \texttt{RoPE} scaling (factor $4.0$).  


Summaries are generated using greedy decoding ($T=0$), capped at $2048$ tokens to ensure fair long-form generation. Each document $d \in D$ is prompted with the following instruction:  
\texttt{"Read the following text and summarize it: \{input document\}. Summarize in \{reference summary word length\} words. Summary:"}  
We deliberately use a generic summarization prompt to evaluate LLMs without giving them any prior indication of which information is salient.\footnote{See Appendix~\ref{app:zs_arg_prompt} for experiments with an argument-aware prompt.} This setup mirrors the human annotation process, where experts also lacked explicit guidance on saliency when producing reference summaries. 
\begin{table}[ht]
\centering
\small
\setlength{\tabcolsep}{4pt}
\renewcommand{\arraystretch}{1.15}

\resizebox{\columnwidth}{!}{
\begin{tabular}{lcccc|cc|cccc}
\toprule
& \multicolumn{4}{c}{CANLII\textsubscript{100}} & \multicolumn{2}{c}{CANLII} & \multicolumn{4}{c}{DRI} \\
\cmidrule(lr){2-5}\cmidrule(lr){6-7}\cmidrule(lr){8-11}
Model & MF & FE & \cellcolor{gray!20}MF & \cellcolor{gray!20}FE & \cellcolor{gray!20}MF & \cellcolor{gray!20}FE & MF & FE & \cellcolor{gray!20}MF & \cellcolor{gray!20}FE \\
\midrule
Qwen-2.5-14B   & \bf{35.7} & 2.0 & \cellcolor{gray!20}\bf{31.4} &  \cellcolor{gray!20} 3.2 & \cellcolor{gray!20}\bf{28.7} & \cellcolor{gray!20}3.1 & \bf{36.5} & 0.2 & \cellcolor{gray!20} \bf{36.7} & \cellcolor{gray!20} 0.5 \\
Qwen-2.5-7B    & \bf{35.5} & 2.0 & \cellcolor{gray!20} \bf{31.2} & \cellcolor{gray!20} 3.6 & \cellcolor{gray!20}\bf{27.8} & \cellcolor{gray!20}3.2 & \bf{33.5} & 0.5 & \cellcolor{gray!20} \bf{34.7} & \cellcolor{gray!20} 1.0 \\
Qwen-2.5-3B    & \bf{36.2} & 2.8 & \cellcolor{gray!20} \bf{30.5} & \cellcolor{gray!20} 3.4 & \cellcolor{gray!20}\bf{29.5} & \cellcolor{gray!20}3.6 & \bf{36.6} & 0.8 & \cellcolor{gray!20} \bf{36.6} & \cellcolor{gray!20} 0.7 \\
Qwen-2.5-1.5B  & \bf{56.3} & 3.6 & \cellcolor{gray!20} \bf{49.1} & \cellcolor{gray!20} 5.5 & \cellcolor{gray!20}\bf{46.9} & \cellcolor{gray!20}4.9 & \bf{49.9} & 0.4 & \cellcolor{gray!20} \bf{47.8} & \cellcolor{gray!20} 1.1 \\
Mistral-8B     & \bf{46.8} & 3.0 & \cellcolor{gray!20} \bf{42.1} & \cellcolor{gray!20} 3.1 & \cellcolor{gray!20}\bf{40.4} & \cellcolor{gray!20}3.7 & \bf{43.3} & 0.4 & \cellcolor{gray!20}\bf{44.0} & \cellcolor{gray!20} 1.3 \\
LLaMA-3.1-8B   & \bf{40.1} & 2.1 & \cellcolor{gray!20} \bf{36.9} & \cellcolor{gray!20} 3.6 & \cellcolor{gray!20}\bf{33.1} & \cellcolor{gray!20}3.3 & \bf{41.1} & 0.3 & \cellcolor{gray!20} \bf{41.7} & \cellcolor{gray!20} 0.8 \\
LLaMA-3.2-3B   & \bf{45.4} & 3.0 & \cellcolor{gray!20} \bf{40.1} & \cellcolor{gray!20} 4.0 & \cellcolor{gray!20}\bf{36.5} & \cellcolor{gray!20}4.2 & \bf{43.8} & 0.3 & \cellcolor{gray!20} \bf{43.3} & \cellcolor{gray!20} 1.3 \\
LLaMA-3.2-1B   & \bf{56.3} & 4.6 & \cellcolor{gray!20} \bf{49.1} & \cellcolor{gray!20} 5.5 & \cellcolor{gray!20}\bf{45.1} & \cellcolor{gray!20}5.7 & \bf{49.9} & 0.7 & \cellcolor{gray!20} \bf{50.0} & \cellcolor{gray!20} 1.3 \\
\bottomrule
\end{tabular}}
\caption{Error rates (\%) normalized by the total number of facts per dataset. Each dataset reports \textbf{Missing Facts (MF)} and \textbf{Factual Errors (FE)}. Columns shaded in gray correspond to \textbf{DeepSeek-R1-Distill-Qwen14B} as the judge. Higher error values are \textbf{bolded}.}
\label{tab:error-rates}
\end{table}
The target length is dynamically matched to the reference summary for comparability.
For \texttt{DRI}, the target length is fixed to the longest reference summary to encourage maximal argument retention.

\begin{figure*}[ht]
\centering
\includegraphics[scale=.3]{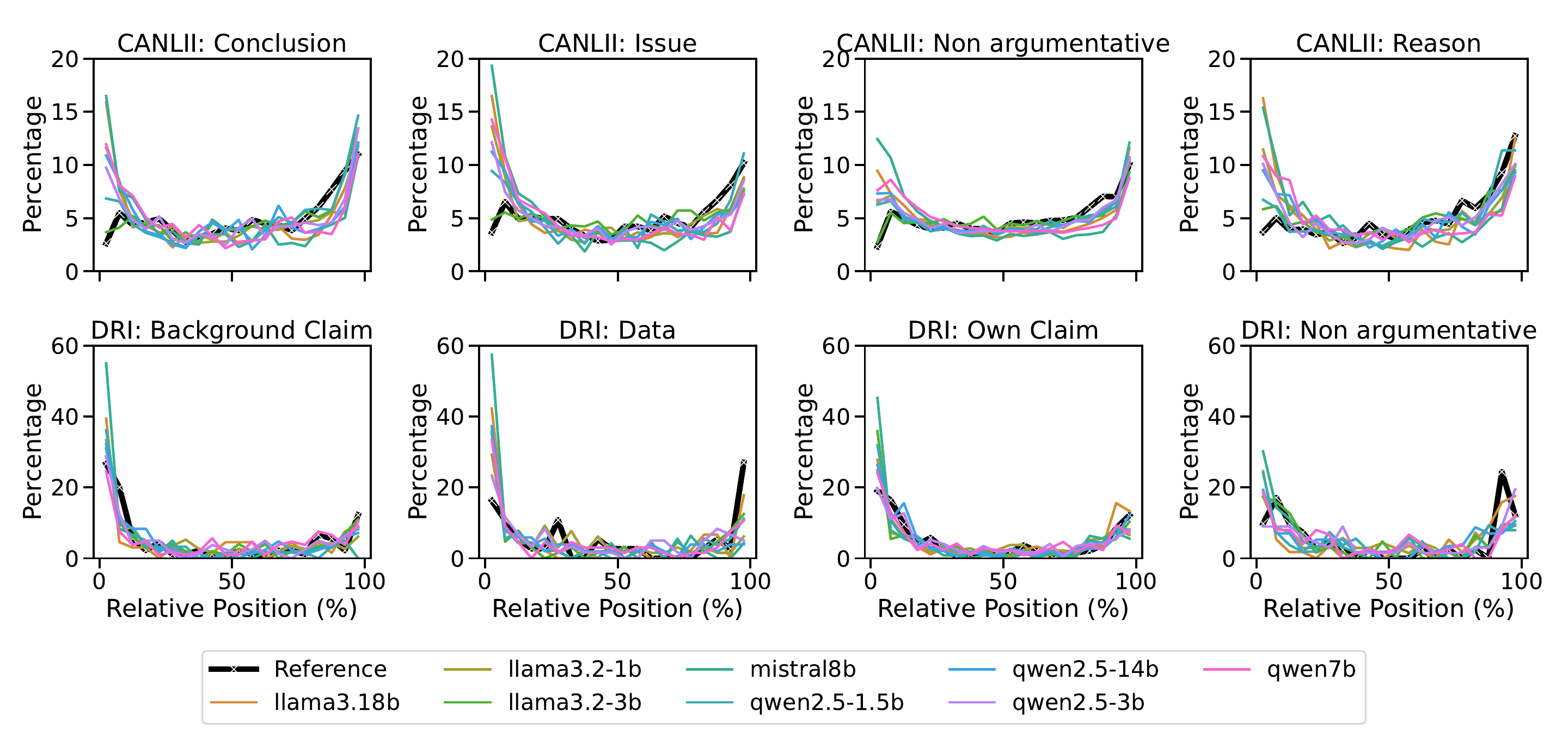}
\caption{Source sentences relative position in the LLM context window across all models and various argument roles for both \texttt{CANLII} and \texttt{DRI} corpora.}
\label{fig:arg_contxt_window}

\end{figure*}


\subsection{RQ1: Can LLM summaries cover salient arguments without explicit supervision?}

We compute ARC\textsubscript{score} for all model-generated summaries described in Section~\ref{subsec:obtain_summs}, 
as well as for human reference summaries on both the \texttt{CANLII} and \texttt{DRI} benchmarks. 
For \texttt{CANLII}, we report results on both the smaller $100$-article subset (\texttt{CANLII\textsubscript{100}}) scored with \texttt{GPT-4o} and the full set (\texttt{CANLII}) scored with \texttt{DeepSeek-R1-Distill-Qwen14B}, 
thereby testing the robustness of ARC\textsubscript{score} across different judges and verifying that it assigns near-perfect scores to expert-written summaries. 
For \texttt{DRI}, because role-level annotations in summaries are unavailable, the goal is to assess how well human references capture salient argument roles compared to LLM-generated outputs.

Table~\ref{tab:arc_scores} shows that argument coverage remains incomplete across all models and domains, indicating that LLMs could still benefit from argument-aware supervision. Although the two judges yield slightly different \texttt{ARC}\textsubscript{score} values, the overall ranking of models is consistent. \footnote{For \texttt{CANLII}\textsubscript{100}, Kendall’s $\tau=0.94$ (between the judges scores); for \texttt{DRI}, $\tau=0.83$; both $p<0.05$.}  
Two clear patterns emerge from the results.  
$(1)$ \textit{Larger models generally achieve higher coverage.} Within the \texttt{Qwen-2.5} family, performance improves when scaling from $1.5$B to $3$B and $7$B across both judges. A similar trend holds for \texttt{LLaMA-3.2}, where the $3$B model surpasses the $1$B variant. \footnote{This scaling effect is examined only for models with $1.5$B–$14$B parameters.}
$(2)$ \textit{Coverage is more challenging in legal opinions.} On \texttt{DRI}, the strongest model (\texttt{Qwen-2.5-7B}) achieves $0.799/0.806$, closely approaching the human reference ($0.838/0.813$) under both judges. By contrast, on \texttt{CANLII}\textsubscript{100} and the full \texttt{CANLII}, the best-performing models plateau at $0.677/0.707$ and $0.725$, far below the human reference ($0.986/0.977$ and $0.982$, respectively). This disparity underscores the difficulty of preserving salient argumentative content in legal texts, where arguments (both salient and non-salient) are sparsely distributed across lengthy contexts.   

\noindent \textbf{Fact-level Analysis.}  
We analyze fact-level decisions (missing versus non-factual) to determine whether coverage errors arise primarily from omissions of key information or from factual inaccuracies. For each model, we compute the proportion in $\%$ of missing facts and factual errors across all generated summaries.  
As shown in Table~\ref{tab:error-rates}, missing facts are the dominant source of error across both datasets, judges, and models, indicating that salient information is more often omitted than misrepresented. While factual inconsistencies do occur, they are consistently less frequent.  
These findings suggest that, beyond hallucination, the central challenge in summarization is achieving comprehensive coverage of salient content.


\subsection{RQ2: Do argument positions 
 in the source affect their coverage in summaries?}
Following \citet{ravaut-etal-2024-context}, we start by analyzing the positions where included LLMs look at in its context window. We leverage the lexical greedy approach for source sentences identification \cite{ravaut-etal-2024-context,adams-etal-2023-sparse} by iteratively adding sentences in the source that maximizes ROUGE-1 score until there is no further improvement.  \footnote{Appendix \ref{app:source_identify} contains the full algorithm.}
We analyze the source sentence indices by their argument role annotations.  
Figure~\ref{fig:arg_contxt_window} shows a clear \texttt{U}-shaped context window bias across all models, most pronounced in \texttt{CANLII}. Argument role analysis indicates that source positions are heavily shaped by this pattern—problematic for \texttt{CANLII}, where reference summaries do not follow fixed positional trends. In contrast, \texttt{DRI} references align more closely with the LLM bias distribution.  

\begin{table}[h]
\centering
\small
\resizebox{\columnwidth}{!}{%
\begin{tabular}{lcc|c|cc}
\toprule
& \multicolumn{2}{c|}{\texttt{CANLII\textsubscript{100}}} & \texttt{CANLII} & \multicolumn{2}{c}{\texttt{DRI}} \\
\cmidrule(lr){2-3}\cmidrule(lr){4-4}\cmidrule(lr){5-6}
\textbf{Model} & $\rho$  & \cellcolor{gray!20}$\rho$  & \cellcolor{gray!20}$\rho$  & $\rho$  & \cellcolor{gray!20}$\rho$  \\
\midrule
Qwen-2.5-14B   & \textcolor{gray}{-0.144} & \cellcolor{gray!20}\textcolor{gray}{-0.131} & \cellcolor{gray!20}-0.146 & \textcolor{gray}{0.119}  & \cellcolor{gray!20}\textcolor{gray}{-0.044} \\
Qwen-2.5-7B    & -0.301                     & \cellcolor{gray!20}-0.237                    & \cellcolor{gray!20}-0.214 & \textcolor{gray}{-0.223} & \cellcolor{gray!20}\textcolor{gray}{0.007} \\
Qwen-2.5-3B    & -0.232                     & \cellcolor{gray!20}\textcolor{gray}{-0.112}  & \cellcolor{gray!20}-0.136 & \textcolor{gray}{0.006}  & \cellcolor{gray!20}\textcolor{gray}{0.039} \\
Qwen-2.5-1.5B  & -0.230                     & \cellcolor{gray!20}\textcolor{gray}{-0.170}  & \cellcolor{gray!20}-0.163 & \textcolor{gray}{0.074}  & \cellcolor{gray!20}\textcolor{gray}{-0.010} \\
Mistral-8B     & -0.369                     & \cellcolor{gray!20}-0.244                    & \cellcolor{gray!20}-0.193 & \textcolor{gray}{-0.055} & \cellcolor{gray!20}\textcolor{gray}{0.012} \\
LLaMA-3.1-8B   & -0.230                     & \cellcolor{gray!20}\textcolor{gray}{-0.088}  & \cellcolor{gray!20}-0.153 & \textcolor{gray}{0.129}  & \cellcolor{gray!20}\textcolor{gray}{0.166} \\
LLaMA-3.2-3B   & -0.216                     & \cellcolor{gray!20}\textcolor{gray}{-0.113}  & \cellcolor{gray!20}-0.190 & \textcolor{gray}{-0.091} & \cellcolor{gray!20}\textcolor{gray}{-0.031} \\
LLaMA-3.2-1B   & \textcolor{gray}{-0.171}   & \cellcolor{gray!20}\textcolor{gray}{-0.087}  & \cellcolor{gray!20}-0.198 & \textcolor{gray}{-0.020} & \cellcolor{gray!20}\textcolor{gray}{-0.031} \\
\bottomrule
\end{tabular}}
\caption{Pearson correlation ($\rho$) between mean relative position of salient arguments and ARC\textsubscript{score}. 
Gray columns = \cellcolor{gray!20}{DeepSeek–R1}, non-shaded = GPT-4o. Values in \textcolor{gray}{gray} are not statistically significant ($p{>}0.05$).}
\label{tab:pearson_correlation_atomic}
\end{table}


\begin{figure*}
  
\includegraphics[scale=.44]{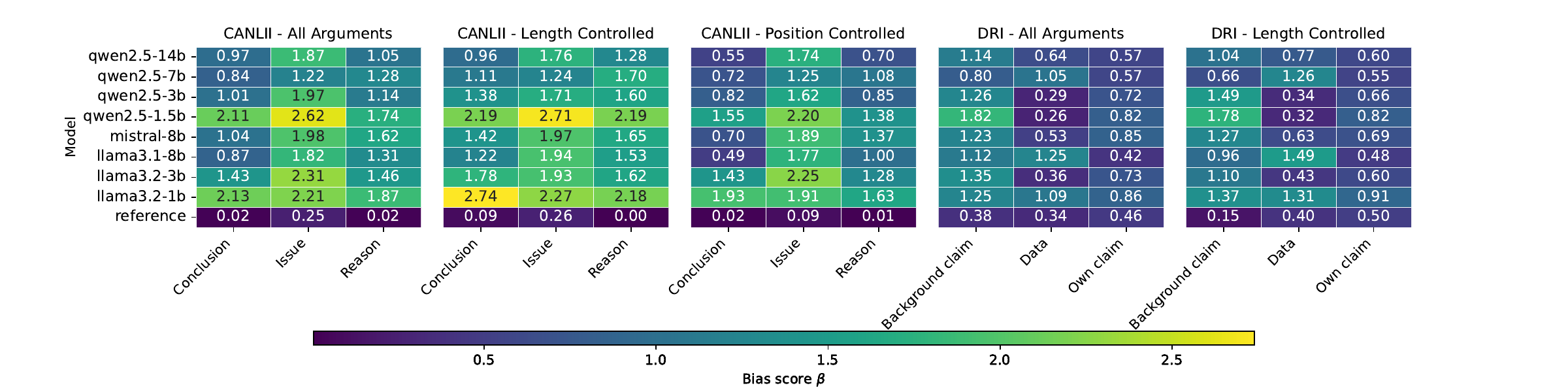}
\caption{\centering Bias score $\beta$ across multiple argument roles for both \texttt{CANLII}\textsubscript{100} and \texttt{DRI} corpora with \texttt{GPT4-o}.}
\label{fig:arg_bias}

\end{figure*}

We analyze how the position of salient arguments affects coverage as measured by $\texttt{ARC}_{\text{score}}$.  
For \texttt{CANLII} and \texttt{CANLII}\textsubscript{100}, we apply greedy sentence selection while restricting the pool to annotated arguments that appear in both the reference summary and the input document. This reduces computational cost and ensures that only arguments included in the reference summary are mapped back to the input.  
For \texttt{DRI}, we directly identify the positions of arguments in the input document with a relevance score of $5$.  
Following \citet{ravaut-etal-2024-context}, we compute the mean relative position of salient arguments and 
measure the Pearson correlation 
between this mean position and $\texttt{ARC}_{\text{score}}$.  

Table~\ref{tab:pearson_correlation_atomic} shows a consistent negative correlation in \texttt{CANLII} (both \texttt{CANLII\textsubscript{100}} with \texttt{GPT4-o} and DeepSeek–R1, as well as the full \texttt{CANLII} set with DeepSeek–R1), with several models reaching significance ($p<0.05$).  
This confirms that LLM context windows systematically bias coverage in long legal cases.  
By contrast, correlations in \texttt{DRI} are weaker and largely non-significant, with some models (e.g., \texttt{LLaMA-3.1-8B})
even showing slight positive trends under DeepSeek–R1.  
These results align with Figure~\ref{fig:arg_contxt_window}: in \texttt{CANLII}, reference summaries diverge from the strong \texttt{U}-shaped bias present in model outputs, whereas in \texttt{DRI}, reference arguments more closely mirror the positional distribution induced by the LLM context window.  



\subsection{RQ3: Are argument roles disproportionately covered?}
\label{subsec:bias_score}

We propose $\beta$ as a role-specific bias score, designed so that higher values indicate stronger bias (i.e., poorer representation of a role).  
The intuition is that if a role $r$ is perfectly represented in the summary, then $\texttt{ARC}_{\text{role}_r} = 1$, and the bias should vanish ($\beta_r = 0$).  
Conversely, lower coverage implies a larger gap from perfect representation, reflected in a higher $\beta_r$.  
Formally, we define:
\[
\beta_r = \left(1 - \texttt{ARC}_{\text{role}_r}\right) \cdot \frac{1}{\log \!\left(1+ \frac{|r|_D}{|\text{args}|_D}\right)} ,
\]
where $|r|_D$ is the frequency of role $r$ in the source document $D$, and $|\text{args}|_D$ is the total number of arguments in $D$.  
The normalization term downweights roles that are overrepresented in the source, ensuring that $\beta_r$ captures true disparities in coverage rather than frequency effects.  

To further reduce confounds from argument length and position—especially in \texttt{CANLII},\footnote{This analysis uses \texttt{CANLII}\textsubscript{100} with the \texttt{GPT4-o} judge, as results in RQ1 and RQ2 show consistent trends across judges and dataset splits.} where longer roles and  arguments positioned in the middle of the document can bias coverage—we compute the bias score $\beta$ within length-controlled groups, allowing a $\pm20\%$ variation in word count. In addition, to control for positional bias effect on coverage, 
we restrict the positional analysis to cases where at least 80\% of arguments fall within either the first or the last 20\% of the document. Positions are computed using the relative positions of the indices of argumentative sentences from the beginning of the document. These constraints help ensure that the observed effects on role coverage are not confounded by argument length or positional bias.

Figure~\ref{fig:arg_bias}\footnote{Appendix~\ref{app:bias_analysis_no_norm} reports results without frequency normalization to rule out denominator inflation effects.} shows that, for \texttt{CANLII}\textsubscript{100}, $\beta$ is consistently lowest for \textit{conclusions}, indicating stronger coverage of this role across both the all-argument analysis and the position-controlled setting. Controlling for role length explains part of the variation across roles, but the relative advantage of \textit{conclusions} persists.  
In \texttt{DRI}, by contrast, bias scores are generally lower than in \texttt{CANLII}\textsubscript{100}, consistent with our earlier finding that LLMs capture more salient arguments in scientific summaries. Nonetheless, \textit{background claims} tend to be covered less effectively, even when length is controlled. Across both domains, model size further influences role-level disparities: for instance, models with  $<3$B parameters consistently exhibit the highest $\beta$, reflecting weaker and less balanced argument coverage.

\section{Conclusion and Future Work}  
We introduced \texttt{ARC}, a hierarchical evaluation framework for assessing how well LLM-generated summaries preserve salient argumentative content. 
\texttt{ARC} provides a principled and interpretable diagnostic tool for evaluating structured argument coverage in long-context summarization. 
Its hierarchical structure not only enables interpretability—revealing which roles are preserved and whether errors arise from omissions or factual inaccuracies—but also achieves higher correlation in holistic evaluation compared to lexical, semantic, entailment, and decomposition-based baselines.
Applying \texttt{ARC} to legal and scientific domains uncovered two consistent limitations of current LLMs: \textit{positional bias}, where the characteristic \texttt{U}-shaped context window negatively affects coverage, and \textit{role bias}, where \textit{conclusions} are favored over other roles such as \textit{issues} and \textit{reasons}. 
Future work can extend this framework by incorporating explicit argument structures into training or prompting, and by leveraging \texttt{ARC}'s interpretable outputs to guide targeted improvements in summarization—particularly in high-stakes domains.


\section*{Limitations}
While the \texttt{ARC} framework enables a comprehensive, multi-level evaluation of argument coverage, several limitations remain that suggest promising directions for future work.  

\noindent \textbf{Limited Benchmarks.}  
Evaluation is constrained by the scarcity of benchmarks explicitly designed for argument coverage. Existing datasets provide limited annotation granularity, especially below the argument-unit level. To our knowledge, only the two benchmarks used here include both salience annotations and explicit argument roles. The small size of \texttt{DRI} (40 documents) also limits generalizability, motivating larger, rigorously annotated corpora—e.g., debates or financial texts—where arguments are central.  

\noindent \textbf{Decomposition Sensitivity.}  
Our decomposition into atomic facts relies on the \citet{yang-etal-2024-fizz} algorithm, LLM-based prompting, and entailment filtering, which may introduce errors. While this decomposition proved more effective than that in \texttt{FactScore}~\cite{min-etal-2023-factscore}, it is worth noting that \texttt{FactScore} relied on human-authored examples from biographical texts, which might have contributed to their lower correlations. Future extensions of \texttt{ARC} should consider incorporating expert-written atomic facts to further improve consistency and balance.

\noindent \textbf{Balancing Precision and Recall.}  
\texttt{ARC} currently emphasizes \textit{recall}—whether salient argument-role facts are preserved—similar to \texttt{FactScore}. While critical for high-stakes domains, this ignores \textit{precision}, i.e., over-inclusion of non-salient content. Future work should jointly assess both dimensions, for instance via a harmonic mean, for a fuller view of coverage quality. \footnote{We additionally conduct a precision--recall analysis on a subset of CANLII summaries; details and results are reported in Appendix~\ref{app:precision_recall}.}

\noindent \textbf{Dependence on Gold Annotations.}  
Our analysis assumes access to gold salience labels. Though this allows conclusive evaluation, future research should explore end-to-end systems that integrate salience detection, fact decomposition, and coverage assessment within a unified framework.

\section*{Ethics Statement}
Our study complies with the ACL Ethics Policy. We primarily evaluate academically available datasets designed explicitly for research purposes,which we obtained through license agreement with the authors of both datasets, thus minimizing privacy risks. Additionally, our work acknowledges potential biases and inaccuracies inherent to LLM-generated outputs, including misrepresentation or omission of critical information from summaries, which could have significant implications in high-stakes domains such as law and science. Researchers and practitioners utilizing our framework should exercise caution and validate results carefully before applying these models in sensitive or consequential decision-making contexts.

\section*{Acknowledgements}
This material is based upon work supported by the National Science Foundation under Grant No. 2040490 and by Amazon.  This research was supported in part by the University of Pittsburgh Center for Research Computing through the resources provided. This research was partially supported by the CS50 fellowship awarded by the department of computer science at the University of Pittsburgh. We want to thank the members of the Pitt PETAL group, Pitt NLP group, and anonymous reviewers for their valuable comments in improving this work.


\bibliography{anthology,custom}
\bibliographystyle{acl_natbib}

\appendix

\section{Extended analysis on included datasets}
\label{app:more_analysis}

\subsection{Examples from included datasets}
Table~\ref{tab:canlii_long_doc_example} presents an excerpt from a legal opinion in the \texttt{CANLII} dataset, with arguments highlighted in both the input and the reference summary. Table~\ref{tab:dri_long_doc_example} provides an excerpt from a scientific article in the \texttt{DRI} dataset, with highlighted arguments in the input. Although the documents are truncated for space, the examples clearly illustrate a key distinction: in \texttt{CANLII}, arguments constitute a smaller fraction of the input, whereas in \texttt{DRI}, the input is densely populated with argumentative content.



\definecolor{issuecolor}{RGB}{0,102,204}    
\definecolor{reasoncolor}{RGB}{0,153,0}      
\definecolor{conclusioncolor}{RGB}{204,0,0}  

\definecolor{issuecolor}{RGB}{0,102,204}    
\definecolor{reasoncolor}{RGB}{0,153,0}      
\definecolor{conclusioncolor}{RGB}{204,0,0}  

\begin{table*}[t]
\centering
\small
\renewcommand{\arraystretch}{1.2}
\setlength{\tabcolsep}{4pt}
\begin{tabular}{p{0.95\textwidth}}
\hline
\textbf{Input Article (Truncated)} \\
\hline
\scriptsize
\textit{
Q.B. A.D. 1987 No. CS 1159 J.C.R., Regina. Applicants seek to quash a search warrant issued by a justice of the peace. 
\textcolor{issuecolor}{\textbf{The respondent, a justice of the peace, issued a search warrant to search a dwelling house for weapons allegedly used in an attempted armed robbery.}} 
The applicants claim the warrant was unlawfully issued without proper grounds.
Specifically, the sworn information relied solely on hearsay from an unidentified informant, lacking corroborating details.
\textcolor{reasoncolor}{\textbf{The applicants argue that no reasonable or probable grounds were disclosed to believe the weapons would be found at the searched location.}} 
They highlight that the informant's reliability was not established, nor was there an oath affirming the informant’s credibility.
The search warrant was issued under Section 443(1)(b) of the Criminal Code, which allows a justice to issue a warrant if reasonable grounds exist to believe evidence of an offence will be found.
\textcolor{reasoncolor}{\textbf{The court explains that on applications to quash a warrant, the reviewing judge cannot substitute their opinion for that of the justice of the peace.}}
Instead, the judge must simply determine whether any evidence existed upon which the justice could be satisfied on reasonable grounds.
Reliance on confidential informants is permitted, even if detailed particulars are absent, provided sufficient basis exists for reliability.
Past cases (e.g., Re Lubell, Re Dodge) have accepted similar levels of disclosure to protect informant anonymity.
\textcolor{reasoncolor}{\textbf{The court notes that substantial compliance with Section 443 is sufficient; perfection in drafting is not required.}}
Given practical constraints faced by peace officers preparing information, reasonable latitude must be given in interpreting the sworn information.
\textcolor{conclusioncolor}{\textbf{The judge concludes that although more information could have been provided, there was sufficient evidence upon which the justice could reasonably issue the warrant.}}
\textcolor{conclusioncolor}{\textbf{Accordingly, the respondent acted within her jurisdiction, and the application to quash the warrant is dismissed.}}
} \\
\hline
\textbf{Reference Summary} \\
\hline
\scriptsize
\textit{
\textcolor{issuecolor}{\textbf{Warrant issued to search a dwelling house for weapons allegedly used in an attempted armed robbery.}} 
The affidavit in support referred to an unknown informant.
\textcolor{reasoncolor}{\textbf{Judge applied the test that the justice of the peace `must be satisfied on reasonable grounds.'}} 
\textcolor{conclusioncolor}{\textbf{Substantial compliance found and warrant upheld.}}
}
\\
\hline
\end{tabular}
\caption{Example of an input legal document (non argumentative text are shortened for space) and its reference summary from \texttt{CANLII}. Highlighted sentences correspond to argumentative roles: \textcolor{issuecolor}{Issue}, \textcolor{reasoncolor}{Reason}, and \textcolor{conclusioncolor}{Conclusion}.}
\label{tab:canlii_long_doc_example}
\end{table*}

\definecolor{ownclaimcolor}{RGB}{128,0,128}    
\definecolor{backgroundcolor}{RGB}{255,140,0}  
\definecolor{datacolor}{RGB}{96,96,96}          

\begin{table*}[t]
\centering
\small
\renewcommand{\arraystretch}{1.2}
\setlength{\tabcolsep}{4pt}
\begin{tabular}{p{0.95\textwidth}}
\hline
\textbf{Input Article (Truncated)} \\
\hline
\scriptsize
\textit{
\textcolor{ownclaimcolor}{\textbf{Our method maintains an explicit polygonal mesh that defines the surface, and an octree data structure that provides both a spatial index for the mesh and a means for efficiently approximating the signed distance to the surface.}} 
\textcolor{backgroundcolor}{\textbf{At each timestep, a new surface is constructed by extracting the zero set of an advected signed-distance function.}}
Semi-Lagrangian backward path tracing is used to advect the signed-distance function.
\textcolor{ownclaimcolor}{\textbf{One of the primary advantages of this formulation is that it enables tracking of surface characteristics, such as color or texture coordinates, at negligible additional cost.}}
\textcolor{datacolor}{\textbf{We include several examples demonstrating that the method can be effectively used as part of a fluid simulation to animate complex and interesting fluid behaviors.}} 
\textcolor{backgroundcolor}{\textbf{The fundamental problem of tracking a surface as it is advected by some velocity field arises frequently in applications such as surface reconstruction, image segmentation, and fluid simulation.}} 
\textcolor{backgroundcolor}{\textbf{Unfortunately, the naive approach of simply advecting the vertices of a polygonal mesh quickly encounters problems such as tangling and self-intersection.}} 
\textcolor{backgroundcolor}{\textbf{Instead, a family of methods, known as level-set methods, has been developed for surface tracking.}}
These methods represent the surface implicitly as the zero set of a scalar field defined over the domain.
\textcolor{backgroundcolor}{\textbf{Level-set methods avoid dealing with topological changes but require high-order conservation law solvers.}}
\textcolor{ownclaimcolor}{\textbf{In contrast, our method constructs a surface directly using semi-Lagrangian contouring without solving PDEs, preserving surface detail efficiently.}}
\textcolor{datacolor}{\textbf{Using adaptive octree data structures, we can efficiently and reliably construct the new surface and corresponding signed-distance function.}}
\textcolor{ownclaimcolor}{\textbf{This allows tracking surface properties such as color or texture coordinates directly on the polygonal mesh during advection, enabling realistic animation of complex fluids.}}
\textcolor{backgroundcolor}{\textbf{Prior methods often suffered from volume loss and smoothing artifacts, particularly in underresolved, high-curvature regions.}}
\textcolor{ownclaimcolor}{\textbf{By using an explicit surface representation, we compute exact distances near the mesh and avoid substantial interpolation errors.}}
...
\textcolor{ownclaimcolor}{\textbf{Finally, the method produces detailed, flicker-free animations of fluid behavior, demonstrating significant advantages over traditional level-set and particle-based approaches.}}
}
\\
\hline
\textbf{Reference Summary} \\
\hline
\scriptsize
\textit{
This article presents a semi-Lagrangian surface tracking method that explicitly represents the surface as a set of polygons.
The new surface and corresponding signed-distance function can be efficiently and reliably constructed using adaptive octree data structures.
One of the primary advantages of this method is that it enables tracking surface characteristics, such as color or texture coordinates, or even simulation variables, accurately at negligible additional cost.
These properties can be easily stored directly on the polygonal mesh and efficiently mapped onto the new surface during semi-Lagrangian advection.
At each timestep, a new surface is constructed by extracting the zero set of an advected signed-distance function.
The explicit representation provides advantages on computing exact signed-distance values near the mesh and storing properties on mesh vertices.
It also facilitates other common operations developed for manipulating and rendering triangle meshes.
To avoid the topological difficulties of directly updating an explicit surface representation, the surface is updated in time through an implicit representation.
The implicit representation is then used to construct a new mesh and extracted using a contouring algorithm.
For its simplicity, robustness, and speed, marching-cubes method is used for contouring.
After the triangle mesh has been extracted, true distance values are assigned to the vertices of octree.
This process is known as redistancing, which comprises three steps: coarsen the octree; compute exact distances at vertices; run a fast marching method over the remaining vertices.
Finally, this method is able to produce detailed, flicker-free animations of complex fluid motions.
}
\\
\hline
\end{tabular}
\caption{Example from \texttt{DRI} showing an input scientific article and its corresponding reference summary. Sentences in the input article are highlighted according to their argument role: \textcolor{ownclaimcolor}{Own Claim}, \textcolor{backgroundcolor}{Background Claim}, \textcolor{datacolor}{Data}. The reference summary is unannotated.}
\label{tab:dri_long_doc_example}
\end{table*}

\subsection{Distribution of arguments across the input}

\begin{figure}[ht!]
    \centering
    \includegraphics[width=0.48\textwidth]{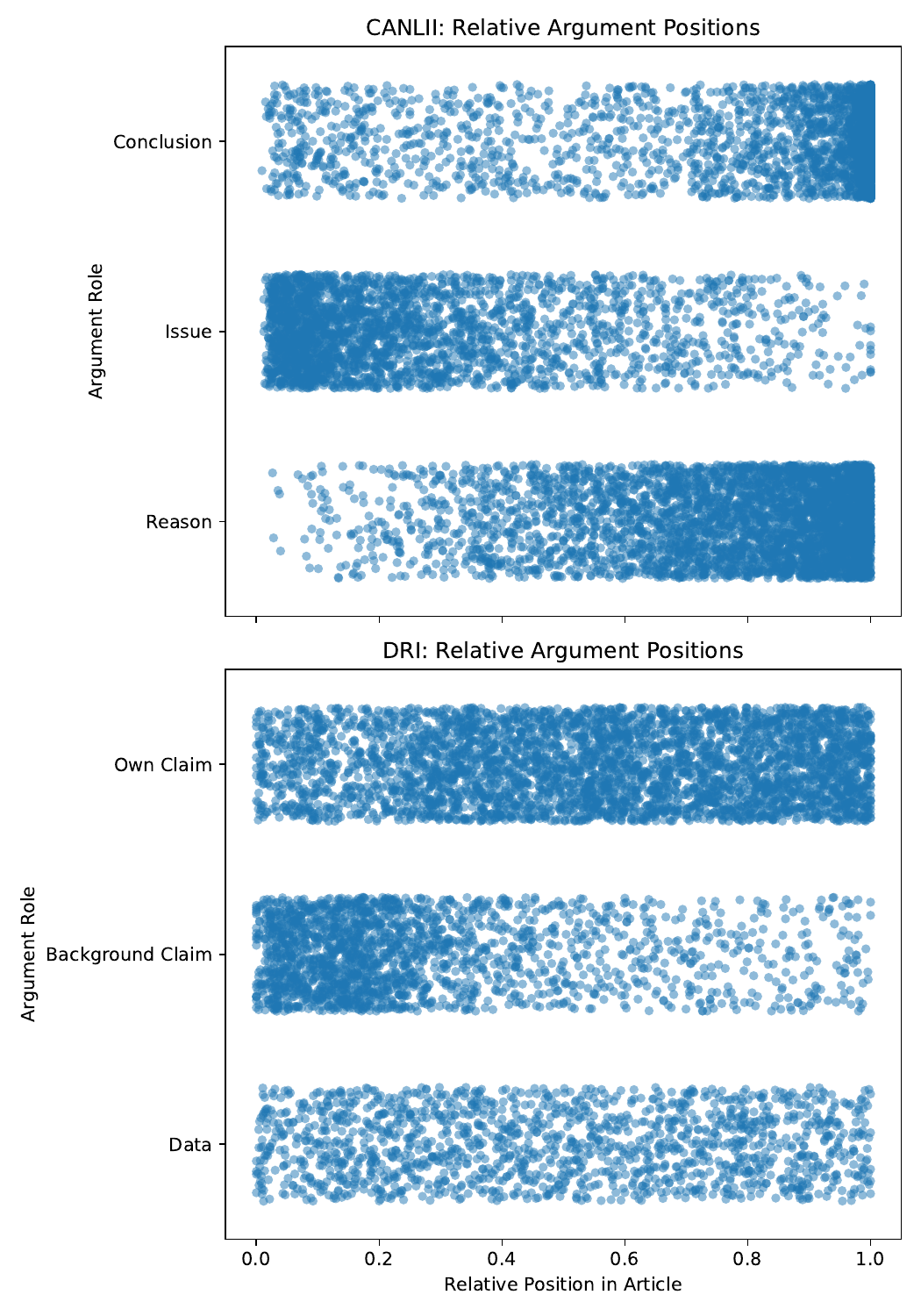}
    \caption{Distribution of argument roles in the input in both \texttt{CANLII} and \texttt{DRI}.}
    \label{fig:dist_arg}
\end{figure}

\begin{figure}[ht!]
    \centering
    \includegraphics[width=0.45\textwidth]{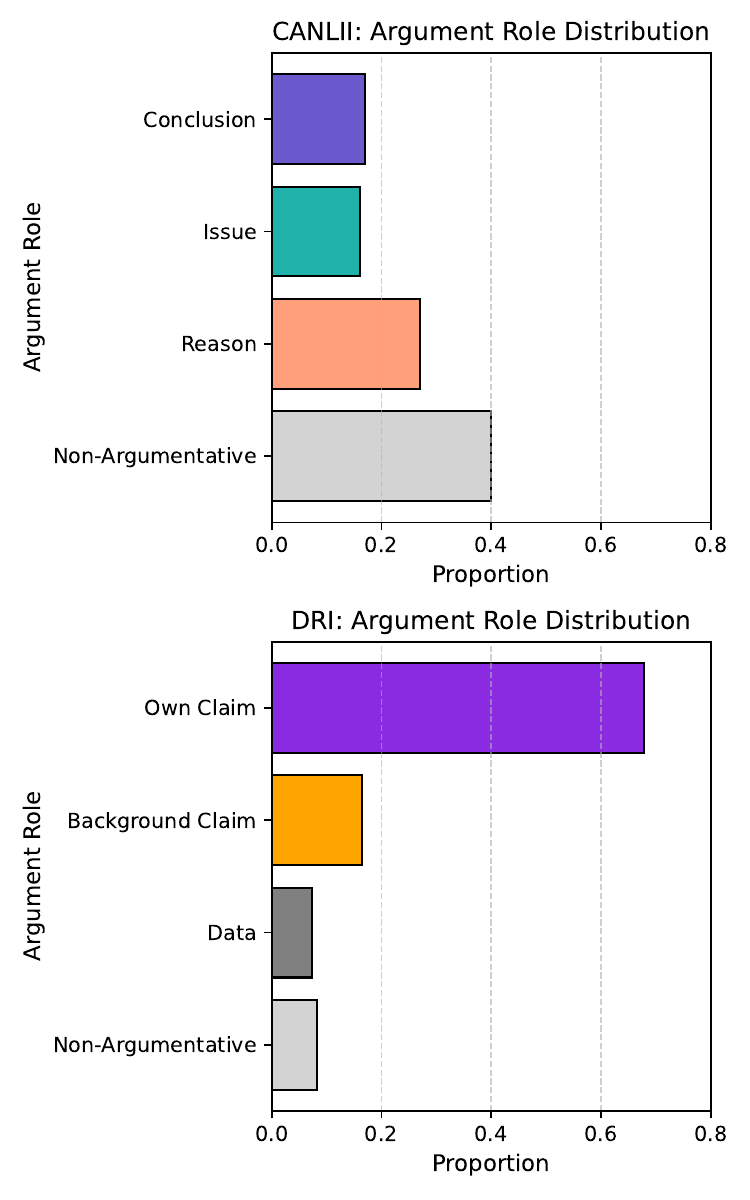}
    \caption{Argument role distributions in summaries for \texttt{CANLII} and \texttt{DRI} (for sentences with relevance score is 5). In \texttt{CANLII}, arguments are less densely represented compared to \texttt{DRI}, where own claims dominate.}
    \label{fig:argument_role_distribution}
\end{figure}

Figure~\ref{fig:dist_arg} illustrates the distribution of argument roles across the source documents. In \texttt{CANLII}, \textit{Conclusion} statements predominantly appear toward the end of the document, while \textit{Issue} statements are concentrated near the beginning. In \texttt{DRI}, \textit{Background claims} are more frequent at the start of the document, which aligns with the conventional structure of scientific writing where literature reviews—typically containing claims from prior work—are introduced early on.

\subsection{Distribution of salient arguments}

Figure~\ref{fig:argument_role_distribution} presents the distribution of argument roles in \texttt{CANLII} reference summaries and in \texttt{DRI} sentences annotated with a relevance Likert score of 5 (indicating very high likelihood of inclusion in a summary). In \texttt{CANLII}, the distribution of argument roles is relatively balanced across categories, whereas in \texttt{DRI}, \textit{own claims}—statements made directly by the authors—dominate the content.

\subsection{Rhetorical roles per relevance to summary Likert score}

\begin{figure}[ht!]
    \centering
    \includegraphics[width=0.5\textwidth]{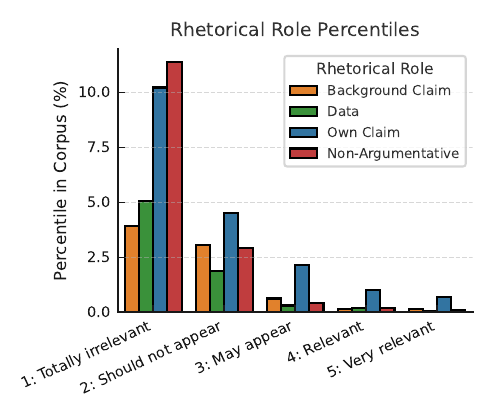}
    \caption{Argument roles per each relevance score to summary from $1$ to $5$.}
    \label{fig:rhet_likert_score}
\end{figure}
To better understand the relationship between rhetorical structure and relevance to the summary, we compute the percentage of each rhetorical role across Likert-rated sentences in the \texttt{DRI} corpus. As shown in Figure~\ref{fig:rhet_likert_score}, non-argumentative content dominates among sentences rated as totally irrelevant to the summary (Likert score 1). However, as the perceived relevance increases, argumentative content becomes more prominent, with Own Claim consistently emerging as the most frequent rhetorical role across all higher-quality categories. This trend highlights a clear shift toward structured argumentative writing in more relevant argument roles.

\section{Fact decomposition algorithm}
\label{app:decompose}

\begin{algorithm}[h!]
\caption{Argument Decomposition and Entailment Filtering}
\label{alg:decomposition_entailment}
\KwInput{Argument $a_i$, Entailment Model $\mathcal{M}$, Entailment Threshold $\tau$}
\KwOutput{Filtered Atomic Facts $\mathcal{F}(a_i)$}

\textbf{Initialization:} \\
$\mathcal{F}(a_i) \gets \emptyset$ \quad (Set of filtered atomic facts) \\

\textbf{Decomposition:} \\
Decompose $a_i$ into atomic facts: $\{m_1, m_2, \dots, m_n\}$ \\

\ForEach{atomic fact $m_j$}{
    Compute entailment using $\mathcal{M}(m_j, a_i) \rightarrow (e, c, n)$ \\
    \If{$e$ (entailment) is predicted}{
        Add $m_j$ to $\mathcal{F}(a_i)$ \\
    }
}

\textbf{Return:} Filtered atomic facts $\mathcal{F}(a_i)$
\end{algorithm}

Algorithm~\ref{alg:decomposition_entailment} outlines the decomposition process for an arbitrary argument \( a_i \in \mathbb{A} \), performed via prompting \texttt{GPT-4o}.

\definecolor{LightGray}{gray}{0.95}

\begin{table*}[!t]
\small
\centering
\renewcommand{\arraystretch}{1.2}
\begin{tabular}{p{0.95\linewidth}}
\toprule
\rowcolor{LightGray}\textbf{Prompt Given to \texttt{GPT4-o} for Argument Decomposition} \\
\midrule
\rowcolor{LightGray}\textbf{Task:} \\
Extract a set of \textbf{atomic facts}—statements that can be \textbf{directly inferred} from the argument without interpretation, assumptions, or redundancy. \\[0.3em]
\rowcolor{LightGray}\textbf{Guidelines:}\\
\begin{itemize}
\item Extract only \textbf{explicitly stated} atomic facts.
\item \textbf{Do not repeat} facts or infer from external knowledge.
\item Maintain \textbf{granularity}: each fact should be \textbf{minimal yet complete}.
\item Output a \textbf{valid Dictionary object} where each key is "fact1", "fact2", etc., and the values are the corresponding atomic facts.
\item \textbf{No additional text or formatting}; dictionary object only.
\item Each argument must yield at least \textbf{one atomic fact}.
\end{itemize} \\[0.3em]
\rowcolor{LightGray}\textbf{Example Output Format:} \\[0.2em]
\texttt{\{ }\\
\texttt{~~~~"fact1": "First atomic fact",} \\
\texttt{~~~~"fact2": "Second atomic fact",} \\
\texttt{~~~~"fact3": "Third atomic fact"} \\
\texttt{\}} \\[0.3em]
\rowcolor{LightGray}\textbf{Input:} \\\{argument\} \\[0.2em]
\rowcolor{LightGray}\textbf{Output:} \\(Dictionary object only) \\
\bottomrule
\end{tabular}
\caption{\centering 
 Prompt provided to \texttt{GPT4-o} for extracting atomic facts from arguments.}
\label{tab:prompt_atomic_decomposition}
\end{table*}

\begin{table*}[!t]
\small
\centering
\renewcommand{\arraystretch}{1.3}
\begin{tabular}{p{0.93\linewidth}}
\toprule
\textbf{Argument (Issue):} \\
FIAT. The father applied to have the mother cited for contempt for denial of access. \\
\midrule
\cmark~ \textbf{Fact 1:} The father applied to have the mother cited for contempt. 
\hfill \textcolor{green!60!black}{(Entailed)} \\

\xmark~ \textbf{Fact 2:} The father applied for denial of access. 
\hfill \textcolor{red!70!black}{(Not-entailed)} \\
\bottomrule
\end{tabular}
\caption{Example of argument decomposition from \texttt{CANLII}, showing atomic facts with entailment status.}
\label{tab:decomposition_example_canlii}
\end{table*}

Table~\ref{tab:prompt_atomic_decomposition} presents the prompt used to extract atomic facts. Table~\ref{tab:decomposition_example_canlii} provides an example decomposition from an \textit{issue} argument role. The second fact is not supported by the original argument and is thus excluded from the final \texttt{ARC} score computation.

\section{Atomic fact error examples}
\label{app:fact_errors}

Table \ref{tab:fact_error_examples_block} presents representative examples from the \texttt{DRI} dataset, illustrating model decisions on atomic facts as evaluated by \texttt{GPT-4o}.
\begin{table*}[t]
\centering
\small
\setlength{\tabcolsep}{6pt}
\renewcommand{\arraystretch}{1.25}
\resizebox{\textwidth}{!}{
\begin{tabular}{p{0.95\textwidth}}
\toprule
\textbf{Example 1 – Non-factual} \\
\midrule
\cellcolor{red!5}
\textbf{Summary:}  
The paper presents an efficient fluid simulation method using Laplacian eigenfunctions as a velocity basis, which avoids the high computational complexity of traditional Eulerian and Lagrangian approaches.  
This method is particularly advantageous for incompressible fluid dynamics, offering a principled way to reduce the dimensionality of the Navier–Stokes equations.  
The authors argue that Laplacian eigenfunctions provide a natural correspondence with spatial scales of vorticity, allowing precise control over turbulent energy at different scales.  
They show that for simple domains, these eigenfunctions have closed-form expressions, enabling fast and exact evaluation of the velocity field.  
The method is formulated using Galerkin projection onto Laplacian eigenfunctions, which are divergence-free and satisfy free-slip boundary conditions.  
For complex domains, a discrete exterior calculus (DEC) approach is used to compute the basis fields.  
The paper highlights the benefits of this method, including its ability to simulate fluid motion interactively, its low memory and precomputation requirements, and its flexibility in controlling viscosity and external forces.  
However, it also acknowledges limitations, such as stair-step artifacts on irregular boundaries and the need for additional methods to handle moving obstacles.  
The authors demonstrate that the method can be applied to various domains, including simple rectangles and complex 3D cavities, and discuss its potential for expressive control and optimization in fluid dynamics.  
The paper concludes by noting that while the method is particularly suited for gaseous phenomena, it may require further development for liquid simulations. \\[4pt]

\textbf{Fact:} Our basis has a natural correspondence with spatial scales of vorticity that is lacking in [Treuille et al. 2006].  

\textbf{Decision:} \cellcolor{red!20}\textbf{Non-factual}  

\textbf{Explanation:}  
The summary correctly states that the basis corresponds to vorticity scales but omits the comparative qualifier “lacking in [Treuille et al. 2006],” misleadingly implying a novel rather than comparative contribution. \\

\midrule
\textbf{Example 2 – Missing Fact} \\
\midrule
\cellcolor{yellow!5}
\textbf{Summary:}  
Pose space deformation (PSD) is a technique that unifies and improves upon both shape interpolation and skeleton-driven deformation methods.  
It represents deformations as mappings from a pose space to displacements in the object's local coordinate frames.  
PSD allows direct manipulation of desired shapes and maintains the performance benefits of traditional methods.  
Key applications include facial and body deformation in entertainment, telepresence, and gaming.  
The limitations of shape interpolation and skeleton subspace deformation (SSD) motivate the development of PSD.  
Shape interpolation lacks smoothness and requires frequent sculpting of new key shapes.  
SSD, while versatile, suffers from the "collapsing elbow" problem and is difficult to control.  
PSD addresses these issues by interpolating deformations in a pose space, enabling smooth and controllable deformations.  
PSD requires scattered data interpolation in high-dimensional spaces.  
Gaussian radial basis functions are used due to their well-behaved nature and ease of implementation.  
The algorithm is simple, general, and allows real-time synthesis, making it suitable for high-resolution models.  
Facial animation can benefit from PSD by allowing sculpting of intermediate expressions and multidimensional pose spaces.  
PSD can handle various deformation scenarios uniformly, from simple joints to complex secondary animations.  
The setup and synthesis costs are minimal, ensuring real-time performance.  
While anatomically based models offer higher quality, PSD provides a practical alternative for real-time applications and fanciful creature designs. \\[4pt]

\textbf{Fact:} The synthesis cost is only slightly more than that of shape interpolation.  

\textbf{Decision:} \cellcolor{yellow!20}\textbf{Missing Fact}  

\textbf{Explanation:}  
The summary omits the comparative detail that PSD’s synthesis cost is slightly higher than that of shape interpolation, overstating the efficiency of the proposed method. \\

\bottomrule
\end{tabular}}
\caption{Full examples of \texttt{ARC}\textsubscript{atomic} factuality and coverage decisions.  
\cellcolor{red!20}Red rows denote \textbf{non-factual} content, and \cellcolor{yellow!20}yellow rows denote \textbf{missing facts}.  
Each example includes the full model-generated summary followed by the evaluated fact, decision, and an explanatory rationale.}
\label{tab:fact_error_examples_block}
\end{table*}

\section{Evaluation prompt}
\label{app:eval_prompts}

Evaluation prompt for fact-level coverage \texttt{ARC}$_{\text{score}}$ is described in Table    \ref{tab:prompt_atomic_evaluation}. We ask the LLM to  generate a rationale before assigning a decision.

\definecolor{LightGray}{gray}{0.95}

\begin{table}[!t]
\small
\centering
\renewcommand{\arraystretch}{1.2}
\begin{tabular}{p{0.95\linewidth}}
\toprule
\rowcolor{LightGray}\textbf{Prompt Given to \texttt{GPT4-o} for Atomic-Level Evaluation} \\
\midrule
\rowcolor{LightGray}\textbf{Task Description:} \\
Given an \textbf{argument} and a \textbf{summary}, evaluate whether the argument is supported by the summary and return a valid tuple in the specified format. \\[0.3em]

\rowcolor{LightGray}\textbf{Explanation:} \\
Provide a brief justification for your decision, identifying any missing, contradictory, or factually incorrect details. \\[0.3em]

\rowcolor{LightGray}\textbf{Return Guidelines:} \\
\begin{itemize}
    \item \textbf{(1, "supported")}: The argument is \textbf{fully supported} by the summary.
    \item \textbf{(0, "missing")}: The argument \textbf{cannot be inferred} from the summary.
    \item \textbf{(0, "not-factual")}: The summary \textbf{contradicts} or misrepresents the argument.
\end{itemize} \\[0.3em]

\rowcolor{LightGray}\textbf{Output Format:} \\
Respond \textbf{only} with a JSON object, structured as: \\[0.2em]
\texttt{\{} \\
\texttt{~~~"explanation": "<explanation placeholder>",} \\
\texttt{~~~"decision": (1, "supported") or (0, "missing") or (0, "not-factual")} \\
\texttt{\}} \\[0.3em]

\rowcolor{LightGray}\textbf{Input:} \\
\texttt{Argument: \{argument\}} \\
\texttt{Summary: \{summary\}} \\[0.3em]

\textbf{Note:} Think critically before deciding. \textbf{Do not generate any extra text beyond the JSON output.} \\
\bottomrule
\end{tabular}
\caption{Prompt provided to \texttt{GPT4-o} for atomic-level argument entailment evaluation.}
\label{tab:prompt_atomic_evaluation}
\end{table}

\section{Likert scores based on human evaluation}
\label{app:human_likert}

Table \ref{tab:final_guidelines} shows the Likert scale from $1$ to $4$ definitions.

\begin{table*}[ht]
\small
\centering
\begin{tabularx}{\textwidth}{lX}   
\toprule
\textbf{Rating} & \textbf{Explanation of the Generated Summary Rating Scale} \\
\midrule
1 & \textbf{No arguments covered:} The generated summary did not cover the highlighted arguments in the reference summary or covered them only inadequately. \\
2 & \textbf{Few arguments covered:} The generated summary adequately covered only a limited number of the highlighted arguments in the reference summary. \\
3 & \textbf{Most arguments covered:} The generated summary adequately covered most of the arguments highlighted in the reference summary. \\
4 & \textbf{All arguments covered:} The generated summary adequately covered all the highlighted arguments in the reference summary. \\
\bottomrule
\end{tabularx}
\caption{Likert scale exact meaning for each score based on definitions obtained from \citet{elaraby-etal-2024-adding}.}
\label{tab:final_guidelines}
\end{table*}

\section{Full expert correlation}
\label{app:exp_corr}

Table~\ref{tab:arc_benchmark_fullexperts} reports correlations with individual experts and their averaged ratings across both judges.
\texttt{ARC}\textsubscript{score} (with \texttt{DeepSeek-R1-Distill-Qwen14B}) achieves the strongest overall correlation with expert judgments, surpassing all baselines.
\texttt{ARC}\textsubscript{score} (with \texttt{GPT4-o}) yields the second-highest overall correlation, outperforming all metrics except \texttt{SummaC}\textsubscript{ZS} (sentence-level) on Expert~2.
These results suggest that the decomposition in \texttt{ARC}\textsubscript{score} not only enables fine-grained and interpretable analyses across error types and argument-role distinctions, but also preserves—and even improves upon—holistic coverage evaluation compared to strong automatic baselines.

\begin{table}[!t]
\small
\centering
\renewcommand{\arraystretch}{1.15}
\setlength{\tabcolsep}{4pt}

\definecolor{arcgpt}{HTML}{E8F4FF}    
\definecolor{arcdsr1}{HTML}{FFF4E5}   

\resizebox{\columnwidth}{!}{%
\begin{tabular}{>{\bfseries}l|cc|cc|cc}
\toprule
\multirow{2}{*}{\textbf{Metric}} & \multicolumn{2}{c|}{\textbf{Expert 1}} & \multicolumn{2}{c|}{\textbf{Expert 2}} & \multicolumn{2}{c}{\textbf{Average}} \\
\cmidrule(lr){2-7}
& $\tau$ & $\rho$ & $\tau$ & $\rho$ & $\tau$ & $\rho$ \\
\midrule
\texttt{ROUGE-1} & 0.378 & 0.520 & 0.347 & 0.456 & 0.391 & 0.539 \\
\texttt{ROUGE-2} & 0.363 & 0.441 & 0.276 & 0.419 & 0.336 & 0.475 \\
\texttt{ROUGE-L} & 0.310 & 0.417 & 0.337 & 0.450 & 0.345 & 0.479 \\
\midrule
\texttt{BERTScore} & 0.319 & 0.409 & 0.292 & 0.398 & 0.354 & 0.517 \\
\midrule
$\texttt{SummaC}_{\text{ZS}}$ (sent) & 0.310 & 0.517 & \it{0.427} & \it{0.517} & 0.387 & 0.537 \\
$\texttt{SummaC}_{\text{ZS}}$ (doc)  & 0.476 & 0.531 & 0.262 & 0.344 & 0.375 & 0.476 \\
$\texttt{SummaC}_{\text{conv}}$ (sent) & 0.329 & 0.371 & 0.341 & 0.460 & 0.345 & 0.459 \\
$\texttt{SummaC}_{\text{conv}}$ (doc)  & 0.408 & 0.300 & 0.269 & 0.264 & 0.352 & 0.311 \\
\midrule
\texttt{FactScore} (GPT4-o) & 0.365 & 0.511 & 0.362 & 0.460 & 0.405 & 0.549 \\
\midrule
\rowcolor{arcgpt}
\texttt{ARC}\textsubscript{score} (GPT-4o) & \it{0.499} & \bfseries \it{0.607} & 0.401 & 0.482 & \it{0.465} & \it{0.593} \\
\rowcolor{arcdsr1}
\texttt{ARC}\textsubscript{score} (DeepSeek-R1) & \bfseries 0.545 & \bfseries 0.653 & \bf{0.441} & \bf{0.519} & \bfseries \bf{0.509} & \bfseries \bf{0.638} \\
\bottomrule
\end{tabular}%
}
\caption{
Metrics correlations (Kendall’s $\tau$, Pearson’s $\rho$) with expert judgments using $87$ articles ($\kappa=0.605$). 
All rows: $p<0.05$. 
Color legend: \colorbox{arcgpt}{\strut\enspace} \texttt{ARC}\textsubscript{score} (\texttt{GPT-4o}), 
\colorbox{arcdsr1}{\strut\enspace} \texttt{ARC}\textsubscript{score} (\texttt{DeepSeek-R1-Qwen-14B}). \textbf{Bolded} is best overall and \textit{ italicized} is second best. 
}
\label{tab:arc_benchmark_fullexperts}
\end{table}

\begin{figure*}
  
\centering
\includegraphics[scale=.43]{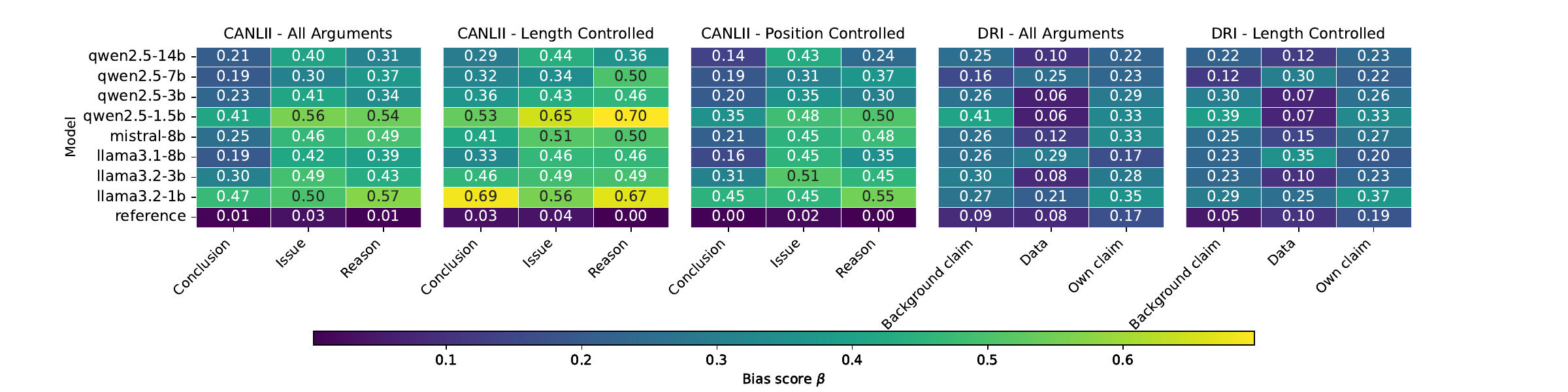}
\caption{Bias score without any frequency normalization ($\beta$=1-\texttt{ARC}\textsubscript{role}) across multiple argument roles (controlled length and non-controlled length) for both \texttt{CANLII}\textsubscript{100} and \texttt{DRI} corpus.}
\label{fig:arg_bias_non_norm}

\end{figure*}

\section{Zero-shot summaries with argument-aware prompts}
\label{app:zs_arg_prompt}
We perform a zero-shot ablation using four representative models from our set of eight: \texttt{Llama-3.2-1B}, \texttt{Llama-3.2-3B}, \texttt{Qwen-2.5-3B}, and \texttt{Qwen-2.5-7B}, varying both model families and sizes.  
In this setting, models are instructed to generate summaries of the same length as the human references, with an explicit emphasis on the key arguments in the input text.  
This experiment evaluates whether explicitly encoding argument salience in the prompt improves the inclusion of salient arguments in model-generated summaries.
The zeroshot prompt used is described in Table \ref{tab:prompt_argument_aware}.




\definecolor{LightGray}{gray}{0.95}

\begin{table}[!t]
\small
\centering
\renewcommand{\arraystretch}{1.2}
\begin{tabular}{p{0.95\linewidth}}
\toprule
\rowcolor{LightGray}\textbf{Argument-Aware Zero-Shot Prompt} \\
\midrule

\rowcolor{LightGray}\textbf{Input:} \\
\textbf{Read the following text carefully:} \\
\texttt{\{input\}} \\[0.3em]

\rowcolor{LightGray}\textbf{Task Description:} \\
 Write an \textbf{abstractive summary} in about \textbf{\{length\}} words. \\[0.3em]

\rowcolor{LightGray}\textbf{Instructions:} \\
\begin{itemize}
    \item Focus primarily on the \textbf{key arguments} presented in the text.
    \item Ensure the summary is \textbf{coherent, fluent, and written in continuous prose} (not bullet points or key phrases).
\end{itemize} \\[0.3em]

\rowcolor{LightGray}\textbf{Output:} \\
\textbf{Summary:} \\

\bottomrule
\end{tabular}
\caption{Argument-aware zero-shot prompt used for abstractive summarization.}
\label{tab:prompt_argument_aware}
\end{table}

Given the consistency across judges and datasets we report results on the smaller \texttt{CANLII} set (\texttt{CANLII}\textsubscript{100}) and the full \texttt{DRI} set using \texttt{GPT-4-o} as a judge.

Table~\ref{tab:arg_prompt_results} shows that explicitly instructing models to focus on key arguments did not consistently improve performance and, in some cases, even degraded coverage of salient roles.
Only \texttt{Qwen-2.5-3B,7B} achieved a modest improvement on \texttt{CANLII}.
These findings suggest that simply prompting models to emphasize arguments is insufficient for enhancing saliency coverage, underscoring the need for deeper alignment or fine-tuning strategies that more effectively encode argumentative relevance.
\begin{table}[t]
\centering
\small
\resizebox{\columnwidth}{!}{
\begin{tabular}{l|cc|cc}
\toprule
\multirow{2}{*}{\textbf{Model}} & \multicolumn{2}{c|}{\texttt{CANLII}} & \multicolumn{2}{c}{\texttt{DRI}} \\
\cmidrule(lr){2-3}\cmidrule(lr){4-5}
 & Vanilla & \cellcolor{gray!10}Arg-Prompt & Vanilla & \cellcolor{gray!10}Arg-Prompt \\
\midrule
LLaMA-3.2-1B & \bf{0.451} & \cellcolor{gray!10}0.426 &\bf{0.674} & \cellcolor{gray!10}0.645 \\
LLaMA-3.2-3B & \bf{0.573} & \cellcolor{gray!10}0.570 & \bf{0.705} & \cellcolor{gray!10}0.667 \\
Qwen-2.5-3B  & 0.648 & \cellcolor{gray!10}\bf{0.654} & \bf{0.730} & \cellcolor{gray!10}0.718 \\
Qwen-2.5-7B  & 0.676 & \cellcolor{gray!10}\bf{0.703} & \bf{0.799} & \cellcolor{gray!10}0.754 \\
\bottomrule
\end{tabular}}
\caption{
Effect of argument-aware prompting on \texttt{ARC}\textsubscript{score} with GPT-4o as the evaluator. Vanilla results are from Table \ref{tab:arc_scores}. \textbf{Bolded} indicates highest across both vanilla and arg-prompted settings.
}
\label{tab:arg_prompt_results}
\end{table}

\section{Source sentences identification}
\label{app:source_identify}
Algorithm~\ref{alg:lexical-greedy-arg} presents the lexical greedy procedure adopted from \citet{ravaut-etal-2024-context} for identifying source sentences given a generated summary. We extend the original algorithm to return both the selected sentence indices and their corresponding argument role types, based on previously annotated sentence-level roles.

\begin{algorithm}[t]
\caption{Lexical Greedy Source Sentence Identification with Argument Roles}
\label{alg:lexical-greedy-arg}
\KwInput{Source document sentences $\mathcal{S}=\{s_1,\dots,s_n\}$, summary $y$, argument role annotations $\mathcal{A}$}
\KwOutput{Selected sentence indices $\mathcal{I}$ and corresponding argument roles $\mathcal{R}$}

$\mathcal{I} \gets \emptyset$\;
$R_{\text{best}} \gets 0$\;

\Repeat{$R_{\text{best}} \le R_{\text{prev}}$}{
    $R_{\text{prev}} \gets R_{\text{best}}$\;
    $i^\ast \gets \texttt{None}$\;

    \ForEach{$s_i \in \mathcal{S}\setminus \mathcal{I}$}{
        $c \gets \textsc{Concat}(\mathcal{I}\cup\{i\})$\;
        $R_i \gets \textsc{Rouge-1}(y, c)$\;
        \If{$R_i > R_{\text{best}}$}{
            $R_{\text{best}} \gets R_i$\;
            $i^\ast \gets i$\;
        }
    }

    \If{$i^\ast \neq \texttt{None}$}{
        $\mathcal{I} \gets \mathcal{I}\cup\{i^\ast\}$\;
    }
}

$\mathcal{R} \gets \{\mathcal{A}(i)\mid i\in \mathcal{I}\}$\;
\Return $\mathcal{I}, \mathcal{R}$\;
\end{algorithm}









\section{Bias analysis for argument coverage without argument role normalization}
\label{app:bias_analysis_no_norm}

While normalization in computing $\beta$ corrects for frequency skew, it may also understate coverage for dominant roles with inherently high raw $\texttt{ARC}_{\text{role}_r}$ scores. To provide a fuller picture, we additionally compute role-specific bias directly as $1-\texttt{ARC}_{\text{role}_r}$. As shown in Figure~\ref{fig:arg_bias_non_norm}, results on \texttt{CANLII}\textsubscript{100} confirm prior findings: LLMs disproportionately prioritize \textit{conclusions} over \textit{issues} and \textit{reasons}, both under length-controlled and non-controlled settings.
For \texttt{DRI}, removing frequency normalization explains partially  the higher $\beta$ $\texttt{ARC}_{\text{role}_{\text{background claim}}}$ scores reported in Section~\ref{subsec:bias_score}.


\section{Precision--recall trade-offs in \texttt{ARC}}
\label{app:precision_recall}

\texttt{ARC} is designed to prioritize \emph{recall}, aligning with prior coverage-based factuality metrics (e.g., \texttt{FactScore}) that emphasize capturing as many salient facts as possible in the summary. Nevertheless, we believe that incorporating precision-based analyses provides a more complete view of different models behavior. To this end, we conducted an additional study on $100$ summaries from the \texttt{CANLII} dataset, evenly sampled across summary length buckets. Using the same atomic fact decomposition  and entailment filtering setup as in the main experiments, we compute: \textbf{Precision}: proportion of covered salient facts divided by the number of facts generated in the summary. \textbf{Recall}: proportion of reference salient facts covered by the summary. \textbf{F1}: harmonic mean of precision and recall. Both atomic fact decomposition and evaluation were conducted using \texttt{GPT4-o}.

Table~\ref{tab:precision_recall_scores} reports precision, recall, and F1 scores across models.


\begin{table}[t]
\centering
\small
\begin{tabular}{lccc}
\toprule
\textbf{Model} & \textbf{Precision} & \textbf{Recall} & \textbf{F1} \\
\midrule
Qwen-2.5-14B   & 0.8766 & 0.6418 & 0.7149 \\
Qwen-2.5-7B    & 0.6337 & 0.6374 & 0.5737 \\
Qwen-2.5-3B    & 0.5027 & 0.5874 & 0.5305 \\
Qwen-2.5-1.5B  & 0.7733 & 0.5608 & 0.6227 \\
Mistral-8B     & 0.7695 & 0.5188 & 0.5724 \\
LLaMA-3.1-8B   & 0.7734 & 0.6174 & 0.6473 \\
LLaMA-3.2-3B   & 0.5622 & 0.3983 & 0.4354 \\
LLaMA-3.2-1B   & 0.4618 & 0.4069 & 0.4080 \\
\bottomrule
\end{tabular}
\caption{Precision, recall, and F1 scores computed using \texttt{ARC}\textsubscript{score} on 100 CANLII summaries (rows ordered by model family and size).}
\label{tab:precision_recall_scores}
\end{table}

To further understand how evaluation choice affects model comparison, Table~\ref{tab:ranking_comparison} shows model rankings induced by each metric.


\begin{table}[t]
\centering
\small
\resizebox{\columnwidth}{!}{\begin{tabular}{lccc}
\toprule
\textbf{Model} & \textbf{Precision Rank} & \textbf{Recall Rank} & \textbf{F1 Rank} \\
\midrule
Qwen-2.5-14B   & 1 & 1 & 1 \\
Qwen-2.5-7B    & 5 & 3 & 5 \\
Qwen-2.5-3B    & 6 & 5 & 4 \\
Qwen-2.5-1.5B  & 3 & 4 & 3 \\
Mistral-8B     & 4 & 6 & 6 \\
LLaMA-3.1-8B   & 2 & 2 & 2 \\
LLaMA-3.2-3B   & 7 & 8 & 7 \\
LLaMA-3.2-1B   & 8 & 7 & 8 \\
\bottomrule
\end{tabular}}
\caption{Model rankings induced by precision, recall, and F1 (rows ordered by model family and size).}
\label{tab:ranking_comparison}
\end{table}

Finally, Table~\ref{tab:rank_correlations} reports rank correlation coefficients between rankings induced by different metrics.

\begin{table}[t]
\centering
\small
\begin{tabular}{lcc}
\toprule
\textbf{Metric Pair} & \textbf{Kendall $\tau$} & \textbf{Pearson $r$} \\
\midrule
Precision vs.\ Recall & 0.50 & 0.62 \\
Precision vs.\ F1     & 0.86 & 0.95 \\
Recall vs.\ F1        & 0.64 & 0.81 \\
\bottomrule
\end{tabular}
\caption{Rank correlations between different evaluation metrics ($p < 0.05$).}
\label{tab:rank_correlations}
\end{table}

Overall, while precision introduces moderate reordering among models, the overall relative ranking of models remains largely stable. The positive correlations between recall and precision indicate that \texttt{ARC}’s recall-oriented design captures a general performance trend that largely persists even under precision-aware evaluation.

\end{document}